\documentclass[conference]{IEEEtran}
\IEEEoverridecommandlockouts
\usepackage{cite}
\usepackage{amsmath,amsthm}
\usepackage{graphicx}
\usepackage{textcomp}
\usepackage{xcolor}
\usepackage{bbm}
\usepackage{dsfont}
\usepackage{subcaption}
\usepackage{url}
\usepackage{hyperref}
\usepackage{cleveref}
\usepackage[font=small]{caption}
\usepackage{algorithm} 
\usepackage{algorithmic}  
\usepackage[ruled,vlined,algo2e]{algorithm2e}

\newcommand{\switch}[3]{\ifthenelse{\equal{#1}{0}}{#2}{#3}}
\newcommand{\arxiv}{1} 

\usepackage[symbol]{footmisc}
\renewcommand{\thefootnote}{\fnsymbol{footnote}}

\usepackage{booktabs,makecell,multirow,hhline,threeparttable} 
\usepackage{xcolor,colortbl,pgf}
\definecolor{Color1}{RGB}{240, 240, 240}
\usepackage{makecell}

\usepackage{dsfont}
\usepackage{amssymb} 
\usepackage{bm} 
\usepackage{xspace}

\newcommand{\tong}[1]{{\color{black}#1}}
\newcommand{\zifeng}[1]{\textcolor{black}{#1}}
\newcommand{\stratis}[1]{\textcolor[rgb]{0,0,0}{#1}}
\newcommand{\edit}[1]{{\color{black}#1}}

\renewcommand{\paragraph}[1]{\noindent\textbf{#1}}

\newcommand{\hb}{HBaR\xspace}
\newcommand{\nol}{L}
\newcommand{\E}{\mathbb{E}}
\newcommand{\reals}{\mathbb{R}}
\newcommand{\naturals}{\mathbb{N}}
\newcommand{\loss}{\mathcal{L}}
\newcommand{\kdloss}{\mathcal{L}_{\mathtt{D}}\xspace}
\newcommand{\celoss}{\mathcal{L}_{\mathtt{CE}}\xspace}
\newcommand{\kl}{\operatorname{KL}}
\newcommand{\aloss}{\tilde{\loss}}
\newcommand{\hbarloss}{\mathcal{L}_{\mathtt{H}}}
\newcommand{\nameloss}{\mathcal{L}_{\mathtt{\name}}}
\newcommand{\admmloss}{\mathcal{L}_{\mathtt{ADMM}}}

\newcommand{\HSIC}{\mathop{\mathrm{HSIC}}}
\newcommand{\name}{PwoA\xspace}

\newcommand{\dx}{d_X}
\newcommand{\dy}{k}

\newcommand{\dzl}{d_{Z_l}}
\newcommand{\aux}{\boldsymbol{\theta}'}
\newcommand{\dual}{\boldsymbol{u}}
\newcommand{\bftheta} {\boldsymbol{\theta}}

\newcommand{\proj}{\mathop{\mathrm{\Pi}}}
\newcommand{\id}{\mathds{1}}

\DeclareMathOperator*{\argmin}{arg\,min}

\def\BibTeX{{\rm B\kern-.05em{\sc i\kern-.025em b}\kern-.08em
    T\kern-.1667em\lower.7ex\hbox{E}\kern-.125emX}}

\newcounter{packednmbr}

\begin{document}

\title{Pruning Adversarially Robust Neural
 Networks without Adversarial Examples}

\author{\IEEEauthorblockN{Tong Jian\textsuperscript{1,$\dagger$}, Zifeng Wang\textsuperscript{1,$\dagger$}, Yanzhi Wang\textsuperscript{2}, Jennifer Dy\textsuperscript{1}, Stratis Ioannidis\textsuperscript{1}}
\IEEEauthorblockA{\textit{Department of Electrical and Computer Engineering} \\
\textit{Northeastern University}\\
\textsuperscript{1}\{jian, zifengwang, jdy, ioannidis\}@ece.neu.edu}
\textsuperscript{2}yanz.wang@northeastern.edu
}

\maketitle
\def\thefootnote{$\dagger$}\footnotetext{Both authors contributed equally to this work.}

\begin{abstract}
\edit{
Adversarial pruning compresses models while preserving robustness. Current methods require access to adversarial examples during pruning. This significantly hampers training efficiency. Moreover, as new adversarial attacks and training methods develop at a rapid rate, adversarial pruning methods need to be modified accordingly to keep up. In this work, we propose a novel framework to prune a previously trained robust neural network while maintaining adversarial robustness, \emph{without} further generating adversarial examples. We 
leverage concurrent self-distillation and pruning to preserve knowledge in the original model as well as regularizing the pruned model via the Hilbert-Schmidt Information Bottleneck. 
We comprehensively evaluate our proposed framework and show its superior performance in terms of both adversarial robustness and efficiency when pruning architectures trained on the MNIST, CIFAR-10, and CIFAR-100 datasets against five state-of-the-art attacks. Code is available at \url{https://github.com/neu-spiral/PwoA/}}.

\end{abstract}


\section{Introduction}





The vulnerability of deep neural networks (DNNs) to adversarial attacks has been the subject of extensive research recently\switch{\arxiv}{~\cite{madry2017towards, cw,autoattack}}{~\cite{moosavi2016deepfool, goodfellow2014explaining, madry2017towards, cw,autoattack}}. Such attacks are intentionally crafted to mislead DNNs towards incorrect predictions, e.g., by adding delicately but visually imperceptible perturbations to original, natural examples~\cite{szegedy2013intriguing}. Adversarial robustness, i.e., the ability of a trained model to maintain its predictive power despite such attacks, is an important property for many safety-critical applications~\cite{adv_self_driving, adv_health, adv_surveillance}. The most common and effective way to attain adversarial robustness is via \emph{adversarial training }~\cite{yan2018deep, zhang2019theoretically, wang2019improving}, i.e., training a model over adversarially generated examples. 
Adversarial training has shown reliable robustness performance against improved attack techniques such as projected gradient descent (PGD)~\cite{madry2017towards}, the Carlini \& Wagner attack (CW)~\cite{cw} and AutoAttack (AA)~\cite{autoattack}. Nevertheless, adversarial training is computationally expensive \cite{madry2017towards,xie2019cvpr}, usually $3\times$--$ 30\times$ \cite{Ali2019nips} longer than natural training, precisely due to the additional cost of generating adversarial examples.

As noted by Madry et al. \cite{madry2017towards}, achieving adversarial robustness requires a significantly wider and larger architecture than that for natural accuracy. The large network capacity required by adversarial training may \stratis{limit its deployment on   resource-constrained hardware or real-time applications.} 
\stratis{Weight} pruning is a prominent compression technique to reduce model size without notable accuracy degradation \switch{\arxiv}{\cite{zhang2018systematic,ren2019admm, Jian2021RFonEdge,wang2022sparcl}.}{ \cite{han2015learning,dong2017learning,liu2018rethinking,zhang2018systematic,ren2019admm, Jian2021RFonEdge,wang2022sparcl}.} While researchers have extensively explored weight pruning, only a few recent works have studied it  jointly with adversarial robustness. Ye et al. \cite{Ye2019advcompression}, Gui et al. \cite{Gui2019unified}, and Sehwag et al. \cite{Sehwag2020hydra} apply active defense techniques with pruning in
their research. However, these works require access to adversarial examples during pruning. Pruning is itself  a laborious process, as effective pruning techniques simultaneously finetune an existing, pre-trained network; incorporating adversarial examples to this process significantly hampers training efficiency. 
\stratis{Moreover, adversarial pruning techniques tailored to specific adversarial training methods need to be continually revised as new  methods develop apace.}

\begin{figure}[!t]
\centering
   \includegraphics[width=1\columnwidth]{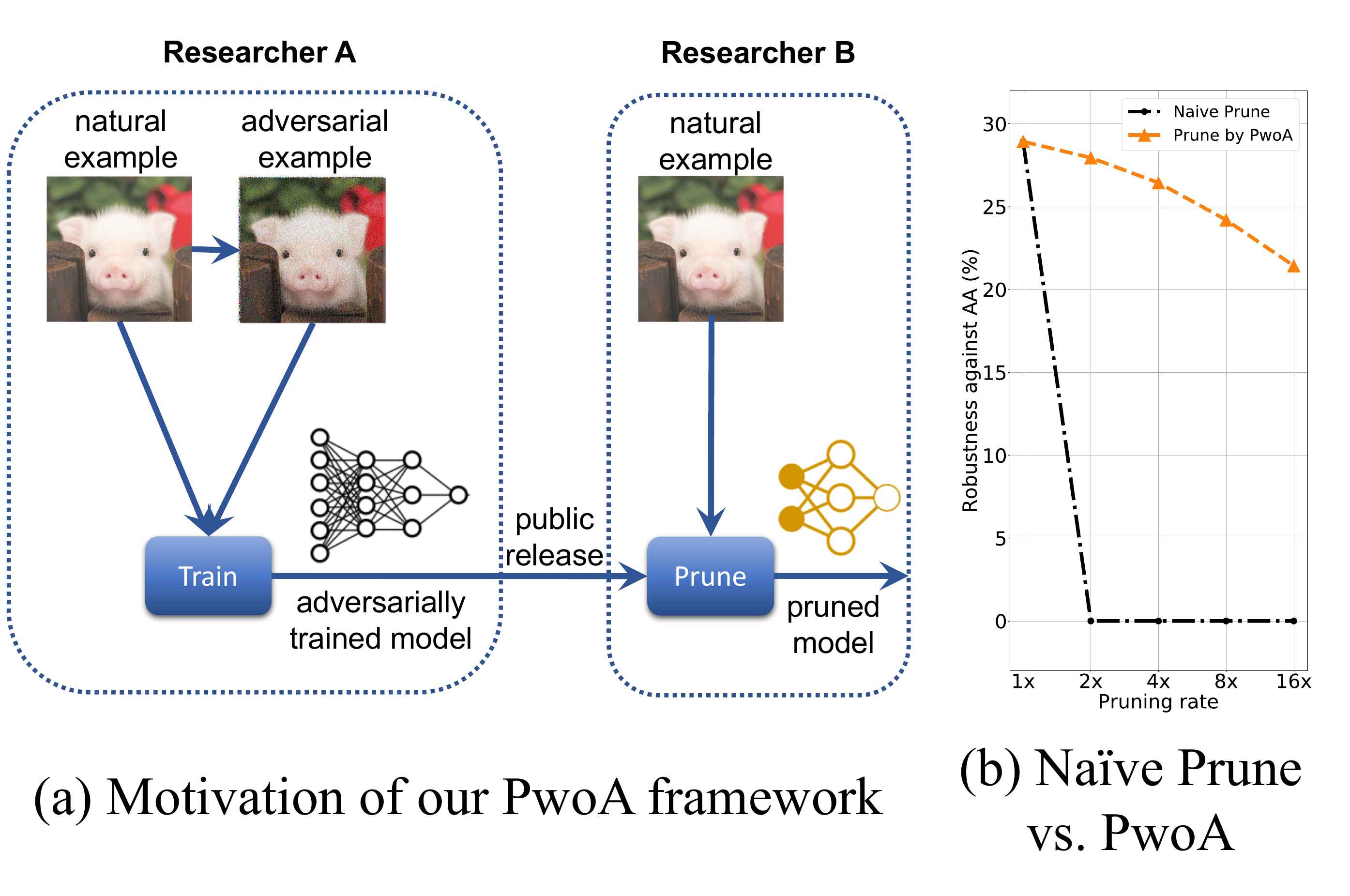}
   \vspace{-25pt}
	\caption{(a) A DNN publicly released by researcher~A,  trained adversarially at a large computational expense, is pruned by Researcher B and made executable on a resource-constrained device. Using \name, pruning by B is efficient, requiring only access to natural examples. (b) Taking a pre-trained WRN34-10 pruned on CIFAR-100 as an example, pruning an adversarially robust model in a na\"ive fashion, without generating any adversarial examples, completely obliterates robustness against AutoAttack \cite{autoattack} even under a $2\times$ pruning ratio. In contrast, our proposed \name framework  efficiently preserves robustness for a broad range of pruning ratios,  without any access to adversarially generated examples. To achieve similar robustness, SOTA adversarial pruning methods  require $4\times$--$7\times$ more training time (see~\Cref{fig:mixed-training} in \Cref{sec:experiments-results}).} 
	\label{fig:overview}
 \vspace{-10pt}
\end{figure}

\stratis{In this paper, we study how take a dense,  adversarially robust DNN, that has already been trained over adversarial examples, and prune it \emph{without any additional adversarial training}.  As a motivating example illustrated in \Cref{fig:overview}(a),  a DNN publicly released by researchers or a company, trained adversarially at a large computational expense, could be subsequently pruned by other researchers to be made executable on a resource-constrained device, like an FPGA. Using our method, the latter could be done efficiently, without access to the computational resources required for adversarial pruning.}

Restricting  pruning to access only natural examples poses a significant challenge. As shown in \Cref{fig:overview}(b), na\"ively pruning a model without  adversarial examples can be catastrophic, obliterating all robustness against AutoAttack. In contrast, our \name is notably robust under a broad range of pruning rates.

Overall, we make the following contributions:
\begin{enumerate}
    \item We propose \name, \zifeng{an end-to-end framework} for pruning a pre-trained adversarially robust model without generating adversarial examples, by \zifeng{(a) \emph{preserving robustness} from the original model via self-distillation \cite{hinton2015distilling, papernot2016distillation, goldblum2020adversarially} and (b) \emph{enhancing robustness} from natural examples via Hilbert-Schmidt independence criterion (HSIC) as a regularizer \cite{ma2020hsic, wang2021hbar}.}
    \item \tong{Our work is the \textit{first to study how an adversarially pre-trained model can be efficiently pruned without access to adversarial examples.} This is an important, novel challenge: prior to our study, it was unclear whether this was even possible. Our approach is generic, and is neither tailored nor restricted to specific pre-trained robust models, architectures, or adversarial training methods.}
    \item We comprehensively evaluate \name on pre-trained adversarially robust models  publicly released by other researchers. In particular, we prune five publicly available  models  that were pre-trained with state-of-the-art (SOTA) adversarial methods on the MNIST, CIFAR-10, and CIFAR-100 datasets. Compared to SOTA adversarial pruning methods, \name can prune a large fraction of weights while attaining comparable--or better--adversarial robustness, at a 4$\times$--7$\times$ training speed up.
\end{enumerate}

\switch{\arxiv}{We omit related work and experimental details from this short paper; both can be found in the extended version \cite{pwoa-extend}.}{
The remainder of this paper is structured as follows. We  review  related work in \Cref{sec:related_work}. In \Cref{sec:background}, we discuss  standard  adversarial robustness, knowledge distillation, and HSIC. In \Cref{sec:methodology}, we present our method. \Cref{sec:experiments} includes our experiments; we conclude in \Cref{sec:conclusions}.}

\section{Related Work} \label{sec:related_work}


\noindent \textbf{Adversarial Robustness.} 
Popular adversarial attack methods include projected gradient descent (PGD) \cite{madry2017towards}, fast gradient sign method (FGSM) \cite{goodfellow2014explaining}, CW attack \cite{cw}, and  AutoAttack (AA)~\cite{autoattack}; see also \cite{adv_review} for a comprehensive review. 
Adversarially robust models are typically obtained via \emph{adversarial training} \cite{pmlr-v97-wang19i}, by augmenting the training set with adversarial examples, generated by the aforementioned adversarial attacks. Madry et al. \cite{madry2017towards} generate adversarial examples via PGD. TRADES \cite{zhang2019theoretically} and MART \cite{wang2019improving} extend adversarial training by incorporating additional penalty terms. LBGAT \cite{cui2021lbgat} guide adversarial training by a natural classifier boundary to improve robustness. However, generating adversarial examples is computationally expensive and time consuming. 

Several recent works observe that information-bottleneck penalties enhance robustness. Fischer \cite{fischer2020conditional} considers a conditional entropy bottleneck (CEB),  while Alemi et al.~\cite{alemi2016deep} suggest a  variational information bottleneck (VIB); both lead to improved robustness properties. Ma et al. \cite{ma2020hsic} and Wang et al. \cite{wang2021hbar} use a penalty based on the Hilbert Schmidt Independence Criterion (HSIC), termed HSIC bottleneck as a regularizer (HBaR). Wang et al. show that HBaR enhances adversarial robustness \emph{even without} generating adversarial examples \cite{wang2021hbar}. For this reason, we incorporate HBaR into our unified robust pruning framework as a means of exploiting adversarial robustness merely from natural examples during the pruning process, without further adversarial training. We are the first to study HBaR under a pruning context; our ablation study (\Cref{sec:experiments-pwoa}) indicates HBaR indeed contributes to enhancing robustness in our setting.  



\noindent \zifeng{\textbf{Adversarial Pruning.}}
Weight pruning is one of the prominent compression techniques to reduce model size with acceptable accuracy degradation. While extensively explored for efficiency and compression purposes \cite{han2015learning,dong2017learning,liu2018rethinking,zhang2018systematic,ren2019admm, Jian2021RFonEdge}, only a few recent works study pruning in the context of adversarial robustness. Several works \cite{Guo2018sparse,kai2018training} theoretically discuss the relationship between adversarial robustness and pruning, but do not provide any active defense techniques. Ye et al. \cite{Ye2019advcompression} and Gui et al. \cite{Gui2019unified} propose AdvPrune to combine the alternating direction method of multipliers (ADMM) pruning framework with adversarial training. 
Lee et al. \cite{lee2021kdprune} propose APD to use knowledge distillation for adversarial pruning optimized by a proximal gradient method. Sehwag et al. \cite{Sehwag2020hydra} propose HYDRA, which uses a robust training objective to learn a sparsity mask. However, all these methods rely on adversarial training. HYDRA further requires training additional sparsity masks, which  hampers training efficiency. In contrast, we distill from a pre-trained adversarially robust model while pruning without generating adversarial examples. Our compressed model can preserve high adversarial robustness with considerable training speedup compared to these methods, as we report in \Cref{sec:experiments-results}.

\section{Background} \label{sec:background}

We use the following standard notation throughout the paper. 
In the standard $\dy$-ary classification setting, we are given a dataset $\mathcal{D} = \{(x_i, y_i)\}_{i= 1}^{n}$, where $x_i \in \reals^{\dx}, y_i \in \{0,1\}^{\dy}$ are i.i.d.~samples drawn from joint distribution $P_{XY}$. Given an $\nol$-layer neural network $h_{\bftheta}:\reals^{\dx}\to\reals^{k}$ parameterized by weights $\bftheta := \{\bftheta_l \}_{l=1}^\nol\in \reals^{d_{\bftheta_l}}$, where ${\bftheta}_l$ is the weight corresponding to the $l$-th layer, for $l=1,\ldots,\nol$, we
define the standard learning objective as follows:
\begin{align}
\loss(\bftheta)= \mathbb{E}_{XY}[\ell(h_{\bftheta}(X), Y)]\approx\frac{1}{n} \sum_{i=1}^n \ell(h_{\bftheta}(x_i), y_i), \label{eq:loss}
\end{align}
where $\ell:\reals^k \times \reals^k \to\reals$ is a loss function,  e.g., cross-entropy.

\subsection{Adversarial Robustness}\label{sec:background-robust}
We call a network \emph{adversarially robust} if it  maintains high prediction accuracy against a constrained adversary that  perturbs input samples. Formally, prior to submitting an input sample  $x\in \reals^{\dx}$, an adversary may perturb $x$ by an arbitrary $\delta\in \mathcal{B}_r$, where $\mathcal{B}_r\subseteq \reals^{\dx}$ is the $\ell_{\infty}$-ball of radius $r$, i.e., \begin{align} \label{def:S_r}
\mathcal{B}_r=B(0,r)=\{\delta \in  \reals^{\dx}: \|\delta\|_{\infty}\leq r\}.\end{align}
The \emph{adversarial robustness} \cite{madry2017towards} of a model $h_{\bftheta}$ is measured by the expected loss attained by such adversarial examples, i.e.,
\begin{equation}
    \label{eq:adv_robustness}
 \begin{split} \aloss(\bftheta)=  \mathbb{E}_{X Y}&\left[\max_{\delta \in \mathcal{B}_r}\ell\left(h_{\bftheta}(X+\delta), Y\right)\right]\\ &\approx  \frac{1}{n} \sum_{i=1}^n \max_{\delta\in \mathcal{B}_r}\ell(h_{\bftheta}(x_i+\delta), y_i).\end{split}
\end{equation}

An adversarially robust neural network $h_{\bftheta}$ can be obtained via \emph{adversarial training}, i.e., by minimizing the adversarial robustness loss in \eqref{eq:adv_robustness} empirically over the training set $\mathcal{D}$. In practice, this amounts to stochastic gradient descent (SGD) over adversarial examples $x_i+\delta$ (see, e.g.,  \cite{madry2017towards}). In each epoch, $\delta$ is generated on a per sample basis via an inner optimization over $\mathcal{B}_r$, e.g., via projected gradient descent (PGD).

Adversarial pruning preserves robustness while pruning. Current approaches combine adversarial training into their pruning objective. In particular, AdvPrune \cite{Ye2019advcompression} directly minimizes adversarial loss $\aloss(\bftheta)$ constrained by sparsity requirements. HYDRA \cite{Sehwag2020hydra} also uses $\aloss(\bftheta)$ to jointly learn a sparsity mask along with $\bftheta_l$. Both are combined with and tailored to specific adversarial training methods, and require considerable training time. This motivates us to propose our \name framework, described in \Cref{sec:methodology}. 

\subsection{Knowledge Distillation}\label{sec:background-kd}
In knowledge distillation \cite{hinton2015distilling, gou2021knowledge}, a student model learns to mimic the output of a teacher. Consider a well-trained teacher model $T$, and a student model $h_{\bftheta}$ that we wish to train to match the teacher's output. Let  $\sigma:\reals^k\to [0,1]^k$ be the softmax function, i.e., $\sigma(\mathbf{z})_j= \frac{e^{z_j}}{\sum_{j'} e^{z_{j'}}}$, $j=1,\ldots,k$. Let 
\begin{align}
   T^\tau(x)&= \sigma \left(  \frac{T(x)}{\tau}\right)~\text{and}~h_{\bftheta}^\tau(x)= \sigma\left(  \frac{h_{\bftheta}(x)}{\tau}\right)
\end{align}
be the softmax outputs of the two models weighed by temperature parameter $\tau>0$ \cite{hinton2015distilling}. Then, the knowledge distillation penalty used to train $\bftheta$ is:
\begin{align} \label{eq:distillation}
 \!\! \mathcal{L}_{\mathtt{KD}}(\bftheta) \!=\! (1\!-\!\lambda)\mathcal{L}(\bftheta) 
    \!+\! \lambda \tau^2 \E_X [ \kl (h_{\bftheta}^\tau(X), T^\tau(X))],\!\!
\end{align}
where $\mathcal{L}$ is the classification loss  of the tempered student network $h_{\bftheta}^\tau$ 
and $\kl$ is the Kullback–Leibler (KL) divergence. 
Intuitively, the knowledge distillation loss $\mathcal{L}_{\mathtt{KD}}$ treats the output of the teacher  as \textit{soft labels} to train the student, so that the student exhibits some inherent properties of the teacher, such as adversarial robustness.

\subsection{Hilbert-Schmidt Independence Criterion}\label{sec:background-hsic}
The Hilbert-Schmidt Independence Criterion (HSIC) is a statistical dependency measure introduced by  Gretton et al.~\cite{gretton2005measuring}. HSIC is the Hilbert-Schmidt norm of the cross-covariance operator between the distributions in Reproducing Kernel Hilbert Space (RKHS). Similar to Mutual Information (MI), HSIC captures non-linear dependencies between random variables. HSIC is defined as:
\begin{equation}    
\begin{split}
\begin{aligned}
\HSIC&(X, Y) 
= \mathbb{E}_{X Y X^{\prime} Y^{\prime}}\left[k_{X}\left(X, X^{\prime}\right) k_{Y^{\prime}}\left(Y, Y^{\prime}\right)\right] \\
&+\mathbb{E}_{X X^{\prime}}\left[k_{X}\left(X, X^{\prime}\right)\right] \mathbb{E}_{Y Y^{\prime}}\left[k_{Y}\left(Y, Y^{\prime}\right)\right] \\
&-2 \mathbb{E}_{X Y}\left[\mathbb{E}_{X^{\prime}}\left[k_{X}\left(X, X^{\prime}\right)\right] \mathbb{E}_{Y^{\prime}}\left[k_{Y}\left(Y, Y^{\prime}\right)\right]\right],
\end{aligned}
\end{split}
\end{equation}
where $X'$ and $Y'$ are independent copies of $X$ and $Y$ respectively, and $k_{X}$ and $k_{Y}$ are kernel functions. 
%
In practice, we often approximate HSIC empirically. Given $n$ i.i.d.~samples $\{(x_i, y_i)\}_{i=1}^{n}$  drawn from $P_{XY}$, 
we estimate HSIC via:
\begin{equation}
    \label{eq:empirical_hsic}
    \widehat{\HSIC}(X, Y)={(n-1)^{-2}} \operatorname{tr}\left(K_{X} H K_{Y} H\right),
\end{equation}
where $K_X$ and $K_Y$ are kernel matrices with entries $K_{X_{ij}}=k_{X}(x_i, x_j)$ and $K_{Y_{ij}}=k_{Y}(y_i, y_j)$, respectively, and $H=I-\frac{1}{n}\id \id ^T$ is a centering matrix.

\section{Problem Formulation}\label{sec:problem-formulate}
Given an adversarially robust model $h_{\bftheta}$, we wish  to efficiently prune non-important weights from this pre-trained model while preserving adversarial robustness of the final pruned model. We minimize the loss function subject to constraints specifying sparsity requirements. More specifically, the weight pruning problem can be formulated as:
\begin{equation}
   \begin{split} \label{prob:prune}
    &\underset{\bftheta}{\text{Minimize:}}\quad \loss(\bftheta),\\
    &\text{subject to\quad} \bftheta_l \in S_{l}, \quad l = 1, \cdots, \nol,
\end{split} 
\end{equation}
where $\loss(\bftheta)$ is the loss function optimizing both the accuracy and the robustness, and $S_l\subseteq \reals^{d_{\bftheta_l}}$ is a weight sparsity constraint set applied to layer $l$, defined as
\begin{equation}\label{eq:irregular}
    S_l = \{ \bftheta_l \mid \|\bftheta_l\|_0 \leq \alpha_l \},
\end{equation}
where $\|\cdot\|_0$ is the size of $\bftheta_l$'s support (i.e., the number of non-zero elements), and $\alpha_l\in \naturals$ is a constant specified as sparsity degree parameter. 


\section{Methodology}\label{sec:methodology}
We now describe \name, our unified framework for pruning a robust network without additional adversarial training. 


\subsection{\zifeng{Robustness-Preserving Pruning}}
\label{sec:prune-kd}
\zifeng{
Given an adversarially pre-trained robust model, we aim to preserve its robustness while sparsifying it via weight pruning. 
In particular, we leverage soft labels generated by the robust model and directly incorporate them into our pruning objective with only access to natural examples.}
Formally, we denote the well pre-trained model by $T$ and its sparse counterpart by $h_{\bftheta}$. The  optimization objective is defined as follows:
\begin{equation}
   \begin{split} \label{prob:prune-kd}
    &\underset{\bftheta}{\text{Min.:}}\quad \kdloss(\bftheta) = \tau^2\E_X [ \kl (h_{\bftheta}^\tau(X), T^\tau(X))],\\
    &\text{subj. to\quad} \bftheta_l \in S_{l}, \quad l = 1, \cdots, \nol,
\end{split} 
\end{equation}
where $\tau$ is the temperature hyperparameter. 
Intuitively, our distillation-based objective forces the sparse model $h_{\bftheta}$ to mimic the soft label produced by the original pre-trained model $T$, while the constraint enforces that the learnt weights are subject to the desired sparsity. This way, we preserve adversarial robustness via distilling knowledge from soft labels efficiently, without regenerating adversarial examples. Departing from the original distillation loss in \eqref{eq:distillation}, we remove the classification loss where labels are used, as we observed that it did not contribute to adversarial robustness \switch{\arxiv}{(see in extended version \cite{pwoa-extend})}{(see \Cref{tab:ce-loss-cifar100} in \Cref{sec:experiments-pwoa})}. 
\zifeng{Solving optimization problem (\ref{prob:prune-kd}) is not straightforward;} we describe how to deal with the combinatorial nature of the sparsity constraints in \Cref{sec:prune-admm}.

\subsection{\zifeng{Enhancing Robustness from Natural Examples}}\label{sec:prune-kd-hbar}

\zifeng{In addition to preserving adversarial robustness from the pre-trained model, we can further enhance robustness directly from natural examples. 
Inspired by the recent work that uses information-bottleneck penalties, \cite{fischer2020conditional,alemi2016deep,ma2020hsic,wang2021hbar}, 
we incorporate  HSIC as a Regularizer (\hb)  into our robust pruning framework. To the best of our knowledge, \hb has only been demonstrated effective under usual adversarial learning scenarios; we are the first to extend it to the context of weight pruning.}
Formally, we denote by $Z_l\in\reals^{\dzl}$, $l\in \{1,\ldots,\nol\}$ the output of the $l$-th layer of $h_{\bftheta}$ under input $X$ 
(i.e., the $l$-th latent representation). The \hb learning penalty \cite{ma2020hsic,wang2021hbar} is defined as follows:
\begin{align}
\label{eq:obj}
\!\hbarloss(\bftheta) =  \!\lambda_{x} \!\sum_{l=1}^{\nol}\! \HSIC(X, Z_l) \!-\!\lambda_{y}\! \sum_{l=1}^{\nol} \!\HSIC(Y, Z_l), \!\!
\end{align}
where $\lambda_{x}, \lambda_{y}\in \reals_+$ are balancing hyperparameters.

Intuitively, since HSIC measures dependence between two random variables, minimizing $\HSIC(X, Z_l)$ corresponds to removing redundant or noisy information from $X$. Hence, this term also naturally reduces the influence of adversarial attack, i.e. perturbation added on the input data. Meanwhile, maximizing $\HSIC(Y, Z_l)$ encourages this lack of sensitivity to the input to happen while retaining the discriminative nature of the classifier, captured by the dependence to useful information w.r.t. the output label $Y$.
\switch{\arxiv}{}{This intrinsic tradeoff  is similar to the so-called information-bottleneck \cite{tishby2000information, tishby2015deep}. Wang et al.~\cite{wang2021hbar} observe this tradeoff between penalties during training; we also observe it during pruning (see Appendix~\ref{sec:appendix-hbar}).}

 \name  combines \hb with self-distillation during weight pruning. We formalize \name to solve the following problem: 
\begin{equation}
   \begin{split} \label{prob:prune-kd-hbar}
    &\underset{\bftheta}{\text{Minimize:}}\quad \nameloss(\bftheta)= \lambda\kdloss(\bftheta)+\hbarloss(\bftheta),\\
    &\text{subject to\quad} \bftheta_l \in S_{l}, \quad l = 1, \cdots, \nol.
\end{split} 
\end{equation}

\subsection{Solving \name via ADMM}\label{sec:prune-admm}
Problem (\ref{prob:prune-kd-hbar}) has combinatorial constraints due to sparsity. Thus, it cannot be solved using stochastic gradient descent  as in the standard CNN training. To deal with this, we follow the ADMM-based pruning strategy by Zhang et al. \cite{zhang2018systematic} and Ren et al. \cite{ren2019admm}. We describe the complete procedure detail in \switch{\arxiv}{\cite{pwoa-extend}.}{Appendix~\ref{sec:appendix-admm}.} In short, ADMM is a primal-dual algorithm designed for constrained optimization problems with decoupled objectives (e.g., problem (\ref{prob:prune-kd-hbar})). Through the definition of an augmented Lagrangian, the algorithm alternates between two primal steps that can be solved efficiently and separately. The first subproblem optimizes objective $\nameloss$ augmented with a proximal penalty; this is an unconstrained optimization solved by classic SGD. The second subproblem is solved by performing Euclidean projections $\proj_{S_{l}}(\cdot)$ to the constraint sets $S_l$; even though the latter are not convex, these projections can be computed in polynomial time. The overall \name framework is summarized in \Cref{alg:pwoa}. 

\begin{algorithm}[!t]
\SetAlgoLined
\textbf{Input:} input samples $\{(x_i, y_i)\}_{i=1}^{n}$, a pre-trained robust neural network $T$ with $L$ layers, mini-batch size $m$, sparsity parameter $\alpha$, learning rate $\beta$, proximal parameters $\{\rho_l\}_{l=1}^L$.\\
\textbf{Output:} parameter of classifier $\theta$\\
 \While{$\theta$ has not converged}{
  Sample a mini-batch of size $m$ from input samples. \\
  SGD step:\\
   $\quad\bftheta \leftarrow \bftheta - \beta \nabla (\nameloss(\bftheta) +\sum_{l=1}^{\nol}\frac{\rho_l}{2}\| \bftheta_l-\aux_{l}+\dual_{l}\|_F^2)$.\\
 Projection step:\\
  $\aux_l \leftarrow \textstyle\proj_{S_{l}}\big(\bftheta_l+\dual_{l}\big),$ for $l=1,\ldots, L.$\\
  Dual variable update step:\\
  $\dual \leftarrow \dual + \bftheta -\aux$
 }
 \caption{\name Framework}
 \label{alg:pwoa}
\end{algorithm}

\section{Experiments}\label{sec:experiments}

\subsection{Experimental Setting}
We conduct our experiments on three benchmark datasets, MNIST, CIFAR-10, and CIFAR-100. To setup adversarially robust pre-trained models for pruning, we consider five adversarially trained models provided by open-source state-of-the-art work, including Wang et al. \cite{wang2021hbar}, Zhang et al. \cite{zhang2019theoretically}, and Cui et al. \cite{cui2021lbgat}, summarized in \Cref{tab:pretrained}.

To understand the impact of each component of \name to robustness, we examine combinations of the following non-adversarial learning objectives for pruning: $\celoss$, $\hbarloss$, and $\kdloss$. All of these objectives are optimized based on natural examples. We also compare \name with \tong{three adversarially pruning methods: APD \cite{lee2021kdprune}}, AdvPrune \cite{Ye2019advcompression} and HYDRA \cite{Sehwag2020hydra}. 

\switch{\arxiv}{}{
\noindent\textbf{Hyperparameters.}
We prune the pre-trained models using SGD  with initial learning rate 0.01, momentum 0.9 and weight decay $10^{-4}$. We set the batch size to 128 for all methods. For our \name, we set the number of pruning and fine-tuning epochs to 50 and 100, respectively. For SOTA methods AdvPrune and HYDRA, we use code provided by authors along with the optimal hyperparameters they suggest. Specifically, for AdvPrune, we set pruning and fine-tuning epochs to 50 and 100, respectively; for HYDRA, we set them to 20 and 100, respectively, and use TRADES as adversarial training loss. We report all other tuning parameters in Appendix~\ref{sec:appendix-params}.

\noindent\textbf{Network Pruning Rate.}
Recall from \Cref{sec:problem-formulate} that the sparsity constraint sets $\{S_l\}_{l=1}^\nol$ are defined by Eq.~\eqref{eq:irregular} with sparsity parameters $\alpha_l\in \naturals$ determining the non-zero elements per layer. We denote the \emph{pruning rate} as the ratio of unpruned size versus pruned size; i.e., for $n_l$ the number of parameters in layer $l$, the pruning rate at layer $l$ can be computed as $\rho_l=\frac{ n_l}{ \alpha_l}$. We set $\alpha_l$ so that we get identical pruning rates per layer, resulting in a uniform pruning rate $\rho$ across the network. 
}

\switch{\arxiv}{\begin{table}[!t]
    \centering
    \setlength{\extrarowheight}{.2em}
    \setlength{\tabcolsep}{2pt}
    \small
    \caption{Summary of the pre-trained models used for datasets.}
    \vspace{2pt}
    \label{tab:pretrained}
    \resizebox{0.48\textwidth}{!}{
    \begin{tabular}{||c ||c c || c c c c  ||}
        \hline
        Dataset & Architecture & Training Method & Natural & PGD$^{20}$ & CW & AA \\
        \hline
        \hline
        \multirow{1}{*}{MNIST}
        & LeNet & PGD \cite{wang2021hbar} & 98.66 & 96.44 & 95.10 & 91.57 \\
        \hline
        \multirow{3}{*}{CIFAR-10}
        & ResNet-18 & TRADES \cite{wang2021hbar} & 84.10 & 52.92 & 51.00 & 49.43 \\
        & WRN34-10 & TRADES \cite{zhang2019theoretically} & 84.96 & 55.44 & 53.92 & 52.34  \\
        & WRN34-10 & LBGAT \cite{cui2021lbgat} & 88.24 & 54.89 & 54.47 & 52.61 \\
        \hline
        \multirow{1}{*}{CIFAR-100}
        & WRN34-10 & LBGAT \cite{cui2021lbgat} & 60.66 & 34.69 & 30.78 & 28.93 \\
        \hline
    \end{tabular}}
\end{table}}{\begin{table*}[!t]
    \centering
    \setlength{\extrarowheight}{.2em}
    \setlength{\tabcolsep}{2pt}
    \small
    \caption{Summary of the pre-trained models used for MNIST, CIFAR-10 and CIFAR-100 datasets.}
    \vspace{2pt}
    \label{tab:pretrained}
    \begin{tabular}{||c ||c c || c c c c c c ||}
        \hline
        Dataset & Architecture & Training Method & Natural & FGSM & PGD$^{10}$ & PGD$^{20}$ & CW & AA \\
        \hline
        \hline
        \multirow{1}{*}{MNIST}
        & LeNet & PGD \cite{wang2021hbar} & 98.66 & 96.02 & 97.53 & 96.44 & 95.10 & 91.57 \\
        \hline
        \multirow{3}{*}{CIFAR-10}
        & ResNet-18 & TRADES \cite{wang2021hbar} & 84.10 & 58.97 & 53.76 & 52.92 & 51.00 & 49.43 \\
        & WRN34-10 & TRADES \cite{zhang2019theoretically} & 84.96 & 60.99 & 56.29 & 55.44 & 53.92 & 52.34  \\
        & WRN34-10 & LBGAT \cite{cui2021lbgat} & 88.24 & 63.62 & 56.34 & 54.89 & 54.47 & 52.61 \\
        \hline
        \multirow{1}{*}{CIFAR-100}
        & WRN34-10 & LBGAT \cite{cui2021lbgat} & 60.66 & 37.46 & 34.99 & 34.69 & 30.78 & 28.93 \\
        \hline
    \end{tabular}
\end{table*}}
\switch{\arxiv}{\begin{table*}[!t]
    \centering
    \setlength{\extrarowheight}{.2em}
    \setlength{\tabcolsep}{2pt}
    \small
    \caption{\textbf{Prune LeNet (PGD), WRN34-10 (LBGAT), and WRN34-10 (LBGAT) on MNIST, CIFAR-10, and CIFAR-100, respectively.} For all the non-adversarial learning objectives, we report natural test accuracy (in \%) and adversarial robustness (in \%) on FGSM, PGD, CW, and AA attacked test examples under different pruning rates. }
    \vspace{2pt}
    \label{tab:overall}
    \resizebox{0.98\textwidth}{!}{
    \begin{tabular}{||c || c c c || cccccc || cccccc || cccccc ||}
        \hline
        \multirow{3}{*}{PR} & \multirow{3}{*}{$\celoss$} & \multirow{3}{*}{$\kdloss$} & \multirow{3}{*}{$\hbarloss$} & \multicolumn{6}{c||}{MNIST} & \multicolumn{6}{c||}{CIFAR-10 } & \multicolumn{6}{c||}{CIFAR-100} \\
        &&&& \multicolumn{6}{c||}{LeNet (PGD)} & \multicolumn{6}{c||}{WRN34-10 (LBGAT)} & \multicolumn{6}{c||}{WRN34-10 (LBGAT)} \\
        &&&& Natural & FGSM & PGD$^{10}$ & PGD$^{20}$ & CW & AA & Natural & FGSM & PGD$^{10}$ & PGD$^{20}$ & CW & AA & Natural & FGSM & PGD$^{10}$ & PGD$^{20}$ & CW & AA \\
        \hline
        \hline
        \multirow{4}{*}{4$\times$}
            & \checkmark &&& \textbf{99.18} & 35.73 & 0.07 & 0.00 & 0.00 & 0.00 & 93.59 & 48.47 & 2.47 & 0.74 & 0.21 & 0.00 & 71.55 & 20.92 & 7.21  & 5.64 & 3.93 & 0.00 \\
            & \checkmark && \checkmark & 98.54 & 91.86 & 89.78 & 78.32 & 79.16 & 47.49 & \textbf{93.68} & 46.52 & 8.45 & 1.69 & 0.25 & 0.00 & \textbf{71.83} & 23.45 & 7.57  & 5.95 & 4.07 & 0.00\\
            &&\checkmark && 98.67 & 95.42 & 97.08 & 95.61 & 95.19 & 89.28 & 88.69 & 62.72 & 52.86 & 50.96 & 50.29 & 48.26 & 60.91 & 36.21 & 32.69 & 31.87 & 27.74 & 25.52 \\
            &&\checkmark & \checkmark & 98.66 & \textbf{95.89} & \textbf{97.35} & \textbf{96.16} & \textbf{96.15} & \textbf{90.00} & 88.51 & \textbf{63.44} & \textbf{53.54} & \textbf{51.51} & \textbf{50.89} & \textbf{49.03} & 60.92 & \textbf{36.70} & \textbf{33.08} & \textbf{32.59} & \textbf{28.40} & \textbf{26.44}  \\
        \hline
        \hline
        \multirow{4}{*}{8$\times$}
            & \checkmark &&& \textbf{99.18} & 39.08 & 0.04  & 0.00 & 0.00 & 0.00 & 93.27 & 41.15 & 0.58 & 0.33 & 0.00 & 0.00 & 71.34 & 15.28 & 3.52  & 2.65 & 1.37 & 0.00 \\
            & \checkmark && \checkmark & 98.63 & 88.70 & 88.89 & 70.67 & 71.42 & 40.71 & \textbf{93.81} & 40.08 & 2.95 & 1.04 & 0.28 & 0.00 & \textbf{71.56} & 17.32 & 3.73  & 2.65 & 1.60 & 0.00\\
            &&\checkmark && 98.66 & \textbf{94.15} & 96.94 & 95.98 & 94.74 & 86.48 & 88.40 & 61.93 & 50.76 & 48.13 & 48.07 & 44.87 & 61.10 & 35.27 & 30.46 & 29.65 & 25.52 & 23.34\\
            &&\checkmark & \checkmark & 98.66 & 95.69 & \textbf{97.13} & \textbf{95.61} & \textbf{95.60} & \textbf{87.37} & 88.66 & \textbf{62.64} & \textbf{51.41} & \textbf{48.98} & \textbf{48.81} & \textbf{46.09} & 61.44 & \textbf{35.61} & \textbf{31.19} & \textbf{30.45} & \textbf{26.32} & \textbf{24.20} \\
        \hline
        \hline
        \multirow{4}{*}{16$\times$} 
            & \checkmark &&& \textbf{98.96} & 79.09 & 0.06  & 0.00 & 0.00 & 0.00 & 92.87 & 20.95 & 0.00 & 0.00 & 0.00 & 0.00 & 69.89 & 14.56 & 3.04  & 2.46 & 1.68 & 0.00\\
            & \checkmark && \checkmark & 98.70 & 81.24 & 83.70 & 50.82 & 54.31 & 13.04 & \textbf{93.14} & 29.88 & 0.84 & 0.11 & 0.04 & 0.00 & \textbf{70.54} & 16.88 & 3.56  & 2.72  & 1.62 & 0.00\\ 
            &&\checkmark && 98.33 & 94.51 & 95.89  & 93.15 & 93.14 & 76.00 & 88.30 & 60.77 & 48.80 & 46.32 & 45.76 & 42.01 & 62.34 & 34.65 & 28.48 & 27.19 & 23.30 & 20.11\\
            &&\checkmark & \checkmark & 98.59 & \textbf{95.03} & \textbf{96.34} & \textbf{94.43} & \textbf{94.48} &\textbf{77.21} & 88.51 & \textbf{61.52} & \textbf{49.68} & \textbf{47.19} & \textbf{47.01} & \textbf{43.33} & 62.53 & \textbf{35.15} & \textbf{29.05} & \textbf{27.88} & \textbf{24.08} & \textbf{21.43}\\
        \hline
    \end{tabular}}
\end{table*}}{\begin{table}[!t]
    \centering
    \setlength{\extrarowheight}{.2em}
    \setlength{\tabcolsep}{2pt}
    \small
    \caption{\textbf{Prune LeNet (PGD) on MNIST.} For all the non-adversarial learning objectives, we report natural test accuracy (in \%) and adversarial robustness (in \%) on FGSM, PGD, CW, and AA attacked test examples under different pruning rates. }
    \vspace{2pt}
    \label{tab:overall-mnist}
    \resizebox{0.48\textwidth}{!}{
    \begin{tabular}{||c || c c c || c c c c c c ||}
        \hline
        PR & $\celoss$ & $\kdloss$ & $\hbarloss$  & Natural & FGSM & PGD$^{10}$ & PGD$^{20}$ & CW & AA \\
        \hline
        \hline
        \multirow{4}{*}{4$\times$}
            & \checkmark &&& \textbf{99.18} & 35.73 & 0.07 & 0.00 & 0.00 & 0.00  \\
            & \checkmark && \checkmark & 98.54 & 91.86 & 89.78 & 78.32 & 79.16 & 47.49 \\
            &&\checkmark && 98.67 & 95.42 & 97.08 & 95.61 & 95.19 & 89.28 \\
            &&\checkmark & \checkmark & 98.66 & \textbf{95.89} & \textbf{97.35} & \textbf{96.16} & \textbf{96.15} & \textbf{90.00} \\
        \hline
        \hline
        \multirow{4}{*}{8$\times$}
            & \checkmark &&& \textbf{99.18} & 39.08 & 0.04  & 0.00 & 0.00 & 0.00  \\
            & \checkmark && \checkmark & 98.63 & 88.70 & 88.89 & 70.67 & 71.42 & 40.71 \\
            &&\checkmark && 98.66 & \textbf{94.15} & 96.94 & 95.98 & 94.74 & 86.48 \\
            &&\checkmark & \checkmark & 98.66 & 95.69 & \textbf{97.13} & \textbf{95.61} & \textbf{95.60} & \textbf{87.37} \\
        \hline
        \hline
        \multirow{4}{*}{16$\times$} 
            & \checkmark &&& \textbf{98.96} & 79.09 & 0.06  & 0.00 & 0.00 & 0.00 \\
            & \checkmark && \checkmark & 98.70 & 81.24 & 83.70 & 50.82 & 54.31 & 13.04 \\ 
            &&\checkmark && 98.33 & 94.51 & 95.89  & 93.15 & 93.14 & 76.00 \\
            &&\checkmark & \checkmark & 98.59 & \textbf{95.03} & \textbf{96.34} & \textbf{94.43} & \textbf{94.48} &\textbf{77.21} \\
        \hline
    \end{tabular}}
\end{table}

\begin{table}[!t]
    \centering
    \setlength{\extrarowheight}{.2em}
    \setlength{\tabcolsep}{2pt}
    \small
    \caption{\textbf{Prune WRN34-10 (LBGAT) on CIFAR-10.} For all the non-adversarial, we report natural test accuracy (in \%) and adversarial robustness (in \%) on FGSM, PGD, CW, and AA attacked test examples under different pruning rates. }
    \vspace{2pt}
    \label{tab:overall-cifar10}
    \resizebox{0.48\textwidth}{!}{
    \begin{tabular}{||c ||c c c || c c c c c c ||}
        \hline
        PR & $\celoss$ & $\kdloss$ & $\hbarloss$  & Natural & FGSM & PGD$^{10}$ & PGD$^{20}$ & CW & AA \\
        \hline
        \hline
        \multirow{4}{*}{4$\times$} 
            & \checkmark &&& 93.59 & 48.47 & 2.47 & 0.74 & 0.21 & 0.00  \\
            & \checkmark && \checkmark & \textbf{93.68} & 46.52 & 8.45 & 1.69 & 0.25 & 0.00 \\
            &&\checkmark && 88.69 & 62.72 & 52.86 & 50.96 & 50.29 & 48.26 \\
            &&\checkmark & \checkmark & 88.51 & \textbf{63.44} & \textbf{53.54} & \textbf{51.51} & \textbf{50.89} & \textbf{49.03} \\
        \hline
        \hline
        \multirow{4}{*}{8$\times$} 
            & \checkmark &&& 93.27 & 41.15 & 0.58 & 0.33 & 0.00 & 0.00 \\
            & \checkmark && \checkmark & \textbf{93.81} & 40.08 & 2.95 & 1.04 & 0.28 & 0.00 \\
            &&\checkmark && 88.40 & 61.93 & 50.76 & 48.13 & 48.07 & 44.87 \\
            &&\checkmark & \checkmark & 88.66 & \textbf{62.64} & \textbf{51.41} & \textbf{48.98} & \textbf{48.81} & \textbf{46.09} \\
        \hline
        \hline
        \multirow{4}{*}{16$\times$} 
            & \checkmark &&& 92.87 & 20.95 & 0.00 & 0.00 & 0.00 & 0.00 \\
            & \checkmark && \checkmark & \textbf{93.14} & 29.88 & 0.84 & 0.11 & 0.04 & 0.00 \\
            &&\checkmark && 88.30 & 60.77 & 48.80 & 46.32 & 45.76 & 42.01 \\
            &&\checkmark & \checkmark & 88.51 & \textbf{61.52} & \textbf{49.68} & \textbf{47.19} & \textbf{47.01} & \textbf{43.33}\\
        \hline
    \end{tabular}}
\end{table}

\begin{table}[!t]
    \centering
    \setlength{\extrarowheight}{.2em}
    \setlength{\tabcolsep}{2pt}
    \small
    \caption{\textbf{Prune WRN34-10 (LBGAT) on CIFAR-100.} For all the non-adversarial, we report natural test accuracy (in \%) and adversarial robustness (in \%) on FGSM, PGD, CW, and AA attacked test examples under different pruning rates.}
    \vspace{2pt} 
    \label{tab:overall-cifar100}
    \resizebox{0.48\textwidth}{!}{
    \begin{tabular}{||c ||c cc|| c c c c c c ||}
        \hline
        PR & $\celoss$ & $\kdloss$ & $\hbarloss$  & Natural & FGSM & PGD$^{10}$ & PGD$^{20}$ & CW & AA \\
        \hline
        \hline
        
        \multirow{4}{*}{4$\times$} 
            & \checkmark &&& 71.55 & 20.92 & 7.21  & 5.64 & 3.93 & 0.00 \\
            & \checkmark && \checkmark & \textbf{71.83} & 23.45 & 7.57  & 5.95 & 4.07 & 0.00 \\
            &&\checkmark && 60.91 & 36.21 & 32.69 & 31.87 & 27.74 & 25.52 \\
            &&\checkmark & \checkmark & 60.92 & \textbf{36.70} & \textbf{33.08} & \textbf{32.59} & \textbf{28.40} & \textbf{26.44} \\
        \hline
        \hline
        \multirow{4}{*}{8$\times$} 
            & \checkmark &&& 71.34 & 15.28 & 3.52  & 2.65 & 1.37 & 0.00 \\
            & \checkmark && \checkmark & \textbf{71.56} & 17.32 & 3.73  & 2.65 & 1.60 & 0.00 \\
            &&\checkmark && 61.10 & 35.27 & 30.46 & 29.65 & 25.52 & 23.34 \\
            &&\checkmark & \checkmark & 61.44 & \textbf{35.61} & \textbf{31.19} & \textbf{30.45} & \textbf{26.32} & \textbf{24.20} \\
        \hline
        \hline
        \multirow{4}{*}{16$\times$} 
            & \checkmark &&& 69.89 & 14.56 & 3.04  & 2.46 & 1.68 & 0.00\\
            & \checkmark && \checkmark & \textbf{70.54} & 16.88 & 3.56  & 2.72  & 1.62 & 0.00 \\
            &&\checkmark && 62.34 & 34.65 & 28.48 & 27.19 & 23.30 & 20.11 \\
            &&\checkmark & \checkmark & 62.53 & \textbf{35.15} & \textbf{29.05} & \textbf{27.88} & \textbf{24.08} & \textbf{21.43}\\
        \hline
    \end{tabular}}
\end{table}}

\noindent\textbf{Performance Metrics and Attacks.}
For all methods, we evaluate the final pruned model via the following metrics. We first measure (a) \emph{Natural} accuracy (i.e., test accuracy over natural examples). We then measure adversarial robustness via test accuracy under (b) \emph{FGSM}, the fast gradient sign attack \cite{goodfellow2014explaining}, (c) \emph{PGD}$^m$, the PGD attack with $m$ steps used for the internal PGD optimization \cite{madry2017towards}, (d) CW (CW-loss within the PGD framework) attack \cite{cw}, and (e) \emph{AA}, AutoAttack \cite{autoattack}, which is the strongest among all four attacks. All five metrics are reported in percent (\%) accuracy. Following prior adversarial learning literature, we set step size to 0.01 and $r= 0.3$ for MNIST, and step size to $2/255$ and $r=8/255$  for CIFAR-10 and CIFAR-100, \stratis{optimizing over $\ell_\infty$-norm balls in all cases}. All attacks happen during the test phase and have full access to model parameters. 
Since there is always a trade-off between natural accuracy and adversarial robustness, we report the best model when it achieves the lowest average loss among the two, as suggested by Ye et al. \cite{Ye2019advcompression} and Zhang et al. \cite{zhang2019theoretically}. We measure and report the overall training time over a Tesla V100 GPU with 32 GB memory and 5120 cores.

\switch{\arxiv}{}{
\begin{table}[!t]
    \centering
    \setlength{\tabcolsep}{2pt}
    \caption{\tong{\textbf{Effect of adding classification loss.} We compare $\nameloss$ (i.e., $\lambda\kdloss+\hbarloss$ where $\lambda=1000$ for CIFAR-100) with $\nameloss+\lambda_{ce}\celoss$ when pruning WRN34-10 (LBGAT) on CIFAR-100. Increasing attention on $\celoss$ improves the natural accuracy while degrading adversarial robustness significantly.}}
    \label{tab:ce-loss-cifar100}
    \resizebox{0.48\textwidth}{!}{
    \begin{tabular}{||c ||ccc| c|| cccccc  ||}
        \hline
        PR & $\celoss$ & $\kdloss$ & $\hbarloss$ & $\lambda_{ce}$ & Natural & FGSM & PGD$^{10}$ & PGD$^{20}$ & CW & AA  \\
        \hline
        \hline
        \multirow{4}{*}{4$\times$} 
            && \checkmark & \checkmark & 0 & 60.92 & 36.70 & \textbf{33.08} & \textbf{32.59} & \textbf{28.40} & \textbf{26.44} \\
        \cline{2-11}
            &\checkmark &\checkmark & \checkmark & 0.01 & 61.74 & 36.87 & 32.15 & 31.61 & 27.34 & 25.31\\
            &\checkmark &\checkmark & \checkmark & 0.1 & 65.03 & 36.03 & 28.98 & 29.02 & 26.41 & 22.52 \\
            &\checkmark &\checkmark & \checkmark & 1 & \textbf{78.66} & \textbf{40.64} & 5.95 & 2.81 & 1.75 & 0.08 \\
        \hline
    \end{tabular}}
\end{table}}

\switch{\arxiv}{}{\begin{figure}[!t]
\begin{tabular}{c c}
   \centering
   \small
   \includegraphics[width=0.45\columnwidth]{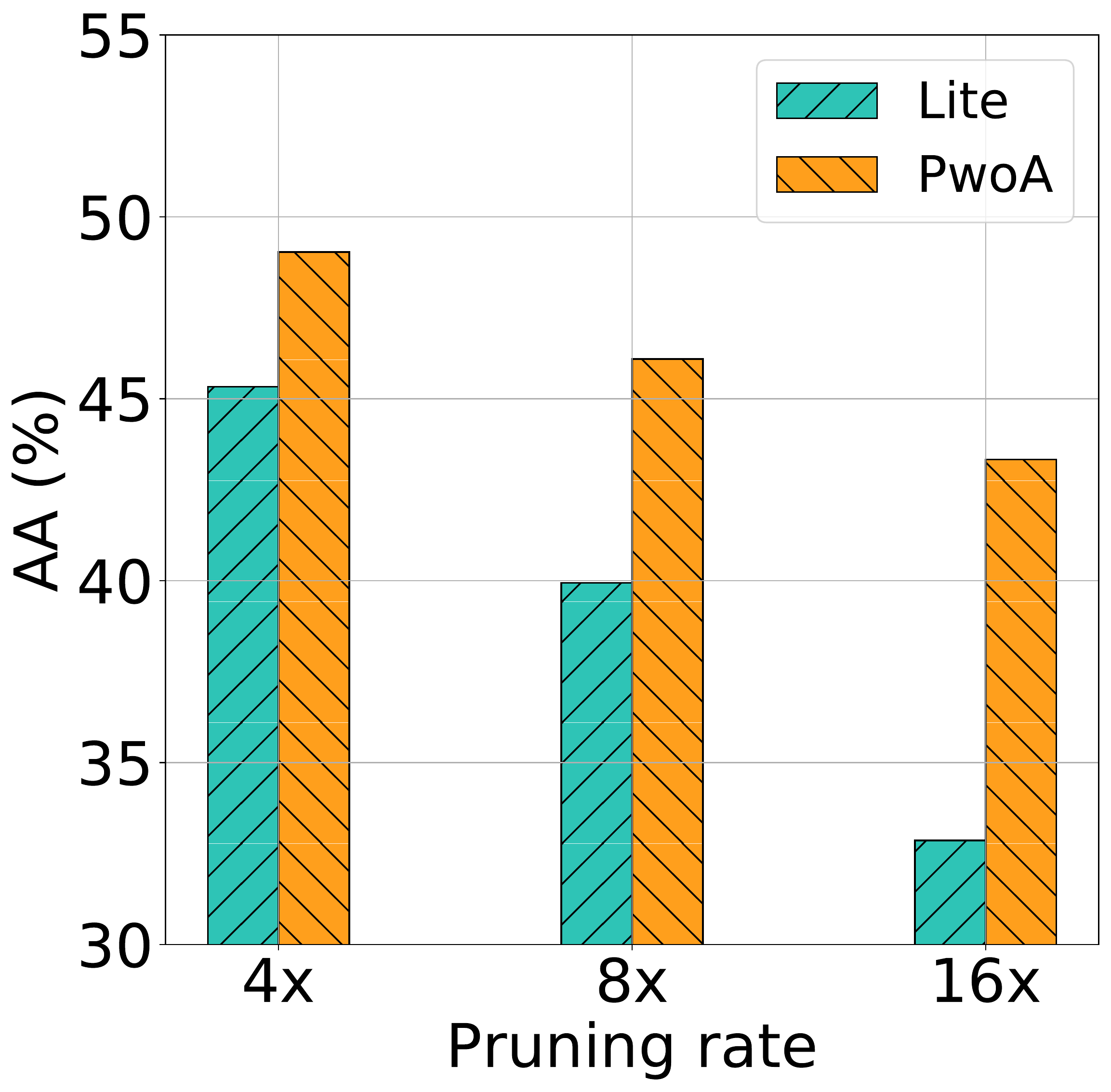} &
   \includegraphics[width=0.45\columnwidth]{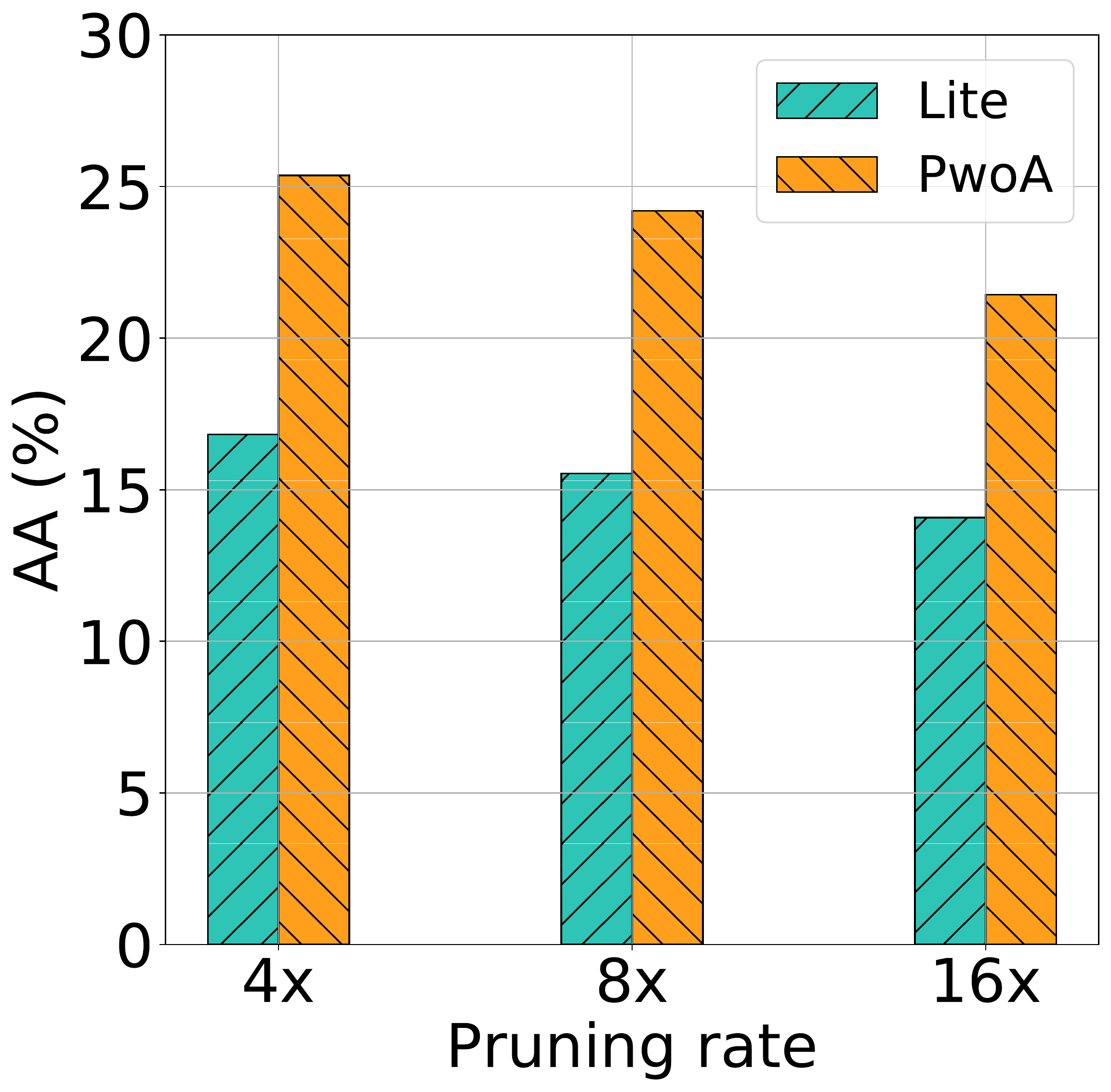} \\
   (a) CIFAR-10 & (b) CIFAR-100 \\
\end{tabular}
\caption{Comparison between the pruned WRN34-10 and adversarially trained WRN34-10-Lite from scratch on (a) CIFAR-10 and (b) CIFAR-100 datasets. Each bar represents the robustness under AA of the corresponding model vs. the pruning rate. The pruned model outperforms its corresponding ‘Lite’ version in all cases, attaining a considerably higher robustness with only access to natural examples.}
\label{fig:pr-woadv}
\end{figure}}

\begin{figure*}[!t]
   \centering
   \small
   \vspace{-5pt}
   \includegraphics[width=2.05\columnwidth]{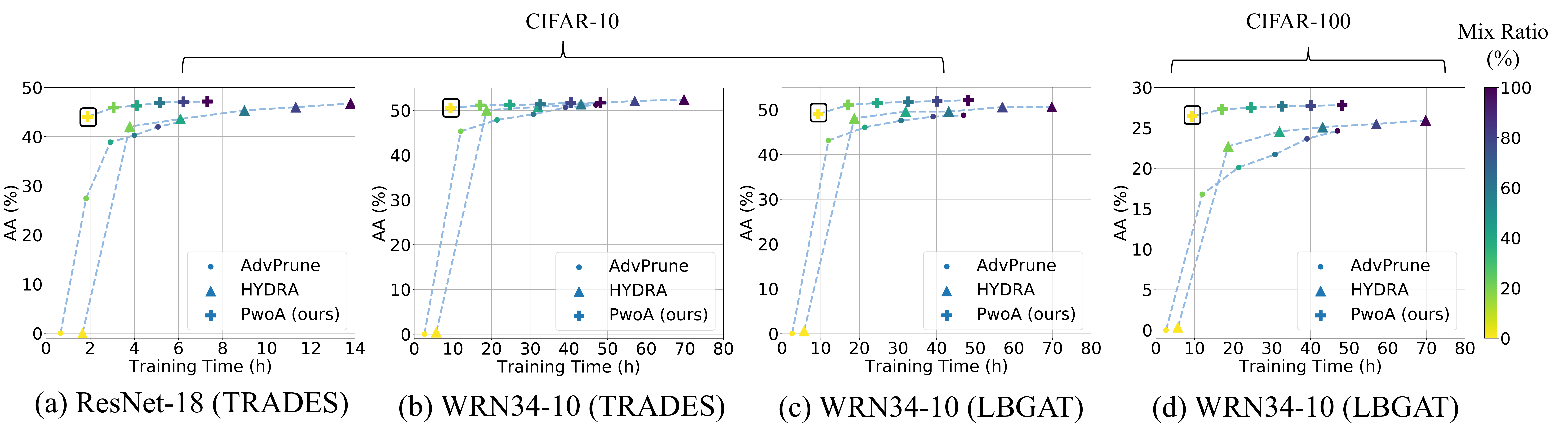} \\
	\caption{Robustness comparison with AdvPrune and HYDRA across different pre-trained models and datasets, under a varying \emph{mix ratio}, i.e.,  fraction (in $\%$) of  natural examples replaced by adversarial examples during training. We plot AA robustness v.s. training time as we modify the mix ratio; \stratis{boxes $\square$ indicate \name with $0\%$ mix ratio (no adversarial examples).} We observe that, competitors are not robust without access to adversarial examples; to achieve \name's robustness  at 0\% mix ratio, AdvPrune and HYDRA require  $4\times$--$7\times$ more training time. On CIFAR-100, they  never meet the performance attained by \name. We also observe that \name improves by partial access to adversarial examples; overall, it attains a much more favorable trade-off between robustness and training efficiency than the two competitors. In fact, in all cases except (b), \name consistently outperforms competitors at 100\% mix ratio, w.r.t.~\emph{both} robustness and training time. }
	\label{fig:mixed-training}
\end{figure*}

\subsection{A Comprehensive Understanding of \name}\label{sec:experiments-pwoa}
\noindent\textbf{Ablation Study and \name Robustness.}
We first examine the synergy between \name terms in the objective in Eq. (\ref{prob:prune-kd-hbar}) and show how these terms  preserve and even improve robustness while pruning. We studied multiple combinations of $\celoss$, $\hbarloss$, and $\kdloss$ in 
\switch{\arxiv}{Table \ref{tab:overall}}{Tables \ref{tab:overall-mnist}-\ref{tab:overall-cifar100}}.
We report the natural test accuracy and adversarial robustness under various attacks of the pruned model under 3 pruning rates (4$\times$, 8$\times$, and 16$\times$) on  MNIST, CIFAR-10, and CIFAR-100. \stratis{For each result reported, we explore hyperparameters $\lambda$ , $\lambda_x$, and $\lambda_y$ as described in \switch{\arxiv}{\cite{pwoa-extend}}{Appendix~\ref{sec:appendix-params}} and report here the best performing values.} 

\emph{Overall, \switch{\arxiv}{Table \ref{tab:overall} suggests}{Tables \ref{tab:overall-mnist}-\ref{tab:overall-cifar100} suggest} that our method \name (namely, $\kdloss+\hbarloss$) prunes a large fraction of weights while attaining the best adversarial robustness for all three datasets}. In contrast, a model pruned by $\celoss$ alone (i.e., with no effort to maintain robustness) catastrophically fails under adversarial attacks on all the datasets. The reason is that when the dataset is more complicated and/or pruning rate is high, $\celoss$ is forced to maintain natural accuracy during pruning, making it deviate from the adversarial robustness of the pre-trained model. In contrast, \emph{concurrent self-distillation ($\kdloss$) and pruning is imperative for preserving substantial robustness without generating adversarial examples during pruning}. We observe this for all three datasets, taking AA under 4$\times$ pruning rate for example, from $0.00\%$ by $\celoss$ to $89.28\%$, $48.26\%$, and $25.52\%$ by $\kdloss$ on MNIST, CIFAR-10 and CIFAR-100, respectively.

We also observe that \emph{incorporating $\hbarloss$ while pruning is beneficial for maintaining high accuracy while improving adversarial robustness against various attacks}. By regularizing $\celoss$ with $\hbarloss$, we observe a sharp adversarial robustness advantage on MNIST, taking AA for example from $0.00\%$ by $\celoss$ to $47.49\%$, $40.71\%$, and $13.04\%$ by incorporating $\hbarloss$ under 4$\times$, 8$\times$, and 16$\times$ pruning rate, respectively; by regularizing $\kdloss$ with $\hbarloss$, we again see that the regularization improves adversarial robustness on all the cases, especially w.r.t. the strongest attack (AA). We note that the robustness improvement of incorporating $\hbarloss$ with $\kdloss$ is not caused by a trade-off between accuracy and robustness: in fact, \emph{$\kdloss+\hbarloss$ consistently improves both natural accuracy and robustness under all pruning rates on all datasets.} Motivated by the above observations, we further analyze how the two terms in \hb defined in Eq. (\ref{eq:obj}) affect natural accuracy and robustness\switch{\arxiv}{; these can be found in the extended version \cite{pwoa-extend}.}{ and summarize these in Appendix~\ref{sec:appendix-hbar}.}

\switch{\arxiv}{}{
\tong{\noindent\textbf{$\celoss$ Diminishes Robustness.}
Recall from \Cref{sec:prune-kd} that we remove the classification loss from the original distillation loss to achieve robustness-preserving pruning. \Cref{tab:ce-loss-cifar100} empirically shows that \textit{classification loss (i.e., $\celoss$) considerably diminishes robustness}. Intuitively, \name distills robustness from the pre-trained robust model rather than acquiring it from natural examples. This is in contrast to observations made with adversarial pruning methods, such as APD \cite{lee2021kdprune}, where the classification loss  increases  robustness. This is because APD prunes by optimizing the original distillation loss \textit{over adversarial examples}, so it may indeed benefit from $\celoss$.
}

\noindent\textbf{Pruning Rate Effect.} On Tables \ref{tab:overall-cifar10}-\ref{tab:overall-cifar100}, we also observe a slight natural accuracy increase during pruning. This is because pruning reduces the complexity of the model, and hence, to some extent, avoids overfitting. However, increasing the pruning rate beyond a critical point can lead to sharp drop in accuracy. This is expected, as reducing the model capacity significantly hampers its expressiveness and starts to introduce bias in predictions. Not surprisingly, this critical point occurs earlier in more complex datasets. We also see that this saturation/performance drop happens earlier for adversarial robustness when compared to natural accuracy: preserving robustness is more challenging, especially without explicitly incorporating adversarial training.}

\switch{\arxiv}{}{\noindent\textbf{Comparison to Na\"ive Parsimony.} We further demonstrate that pruning while training is imperative for attaining high robustness under a parsimonious model. To show this, we construct a class of models that has fewer parameters than the original WRN34-10, and explore the resulting robustness-compression trade-off. We term the first class of models as ‘WRN34-10-Lite’: these models have the same architecture WRN34-10 but contain fewer filters in each convolutional layer (resulting in fewer parameters in total). These WRN34-10-Lite models are   designed to have similar total number of parameters as pruned models with pruning rates 4$\times$, 8$\times$, and 16$\times$, respectively. We train these ‘Lite’ models for 100 epochs on adversarial examples generated by PGD$^{10}$. The pruned model outperforms its corresponding ‘Lite’ version in all cases, improving robustness under 16$\times$ against AA by 10.47\% and 7.35\%, on CIFAR-10 and CIFAR-100, respectively.}

\subsection{Comparison to Adversarial Pruning (AP) Methods}\label{sec:experiments-results}

\noindent\textbf{Robustness with Partial Access to Adversarial Examples.}
\tong{We\switch{\arxiv}{\noindent}{ first} compare \name with two state-of-the art AP baselines, i.e., AdvPrune and HYDRA,} in terms of adversarial robustness and training efficiency on the CIFAR-10 and CIFAR-100 datasets. Both AdvPrune and HYDRA require access to adversarial examples. To make a fair comparison, we generate adversarial examples progressively for all methods, including \name: in \Cref{fig:mixed-training}, we change the \emph{mix ratio}, i.e., the fraction of total natural examples replaced by adversarial examples generated by PGD$^{10}$. We plot AA robustness vs. training time, under a $4\times$ pruning rate.  We observe that, without access to adversarial examples (mix ratio 0\%), both competing methods fail catastrofically, exhibiting no robustness whatsoever. Moreover, to achieve the same robustness as \name, they require between \textbf{$\mathbf{4\times}$ and $\mathbf{7\times}$} more training time; on CIFAR-100, they actually never meet the performance attained by \name. We also observe that \name improves by partial access to adversarial examples; 
overall, it attains a much more favorable trade-off between robustness and training efficiency than the two competitors. Interestingly, with the exception of the case shown in \Cref{fig:mixed-training}(b) (WRN34-10 over CIFAR-10), \name consistently outperforms competitors at 100\% mix ratio, w.r.t.~\emph{both} robustness and training time.


\noindent\textbf{Impact of Pre-training Method.} We also observe that HYDRA performs well when pruning models pre-trained with TRADES, but gets worse when dealing with model pre-trained with LBGAT. This is because HYDRA prunes the model using TRADES as adversarial loss, and is thus tailored to such pre-training. When models are pre-trained via LBGAT, this change of loss hampers performance. In contrast, \name can successfully prune an arbitrary pre-trained model, irrespective of the architecture or pre-training method.

\switch{\arxiv}{}{\begin{table}[!t]
    \centering
    \setlength{\extrarowheight}{.2em}
    \setlength{\tabcolsep}{2pt}
    \small
    \caption{\textbf{Prune WRN34-10 (LBGAT) on CIFAR-10}: Comparison of \name with SOTA methods w.r.t various attacks and training time (TT, in \textit{h}) under different pruning rates at 20\% mix ratio.}
    \vspace{2pt}
    \label{tab:compare-cifar10}
    \resizebox{0.48\textwidth}{!}{
    \begin{tabular}{||c |c || c c c c c c| c |}
        \hline
        PR & Methods & Natural & FGSM & PGD$^{10}$ & PGD$^{20}$ & CW & AA & TT \\
        \hline
        \hline
        \multirow{3}{*}{4$\times$}
            & AdvPrune    & \textbf{89.35} & 58.05 & 47.14 & 45.01 & 45.68 & 43.31 & 12.01 \\
            & HYDRA       & 86.07 & 57.45 & 51.30 & 50.20 & 50.01 & 48.09 & 18.77 \\
            & \name (ours)& 88.10 & \textbf{62.96} & \textbf{55.40} & \textbf{53.72} & \textbf{53.30} & \textbf{51.07} & 17.06 \\
        \hline
        \hline
        \multirow{3}{*}{8$\times$}
            & AdvPrune    & \textbf{89.31} & 57.91 & 47.18 & 45.22 & 45.45 & 43.20 & 12.44 \\
            & HYDRA       & 86.50 & 57.89 & 51.28 & 50.20 & 50.15 & 48.09 & 18.89 \\
            & \name (ours)& 88.11 & \textbf{62.86} & \textbf{54.64} & \textbf{52.93} & \textbf{52.48} & \textbf{50.07}& 17.13 \\
        \hline
        \hline
        \multirow{3}{*}{16$\times$} 
            & AdvPrune    & \textbf{89.37} & 55.32 & 46.68 & 44.77 & 44.13 & 42.61 & 12.38 \\
            & HYDRA       & 85.98 & 57.38 & 51.17 & 50.27 & 49.34 & 47.74 & 18.91 \\
            & \name (ours)& 88.10 & \textbf{62.04} & \textbf{53.38} & \textbf{51.35} & \textbf{51.03} & \textbf{48.44}& 17.11 \\
        \hline
    \end{tabular}}
\end{table}}

\begin{table}[!t]
    \centering
    \setlength{\extrarowheight}{.2em}
    \setlength{\tabcolsep}{2pt}
    \small  
    \caption{\textbf{Prune WRN34-10 (LBGAT) on CIFAR-100}: Comparison of \name with SOTA methods w.r.t various attacks and training time (TT, in \textit{h}) under different pruning rates at 20\% mix ratio.}
    \vspace{2pt}
    \label{tab:compare-cifar100}
    \resizebox{0.48\textwidth}{!}{
    \begin{tabular}{||c |c || c c c c c c | c||}
        \hline
        PR & Methods & Natural & FGSM & PGD$^{10}$ & PGD$^{20}$ & CW & AA & TT \\
        \hline
        \hline
        \multirow{3}{*}{4$\times$}
            & AdvPrune    & \textbf{68.39} & 40.77 & 24.71 & 22.42 & 21.45 & 14.95 & 12.14\\
            & HYDRA       & 60.61 & 29.54 & 25.88 & 25.21 & 24.22 & 22.81 & 18.69 \\
            & \name (ours)& 60.93 & \textbf{36.92} & \textbf{33.62} & \textbf{33.30} & \textbf{29.10} & \textbf{27.31}& 17.03 \\
        \hline
        \hline
        \multirow{3}{*}{8$\times$}
            & AdvPrune    & \textbf{68.33} & 40.73 & 24.34 & 22.03 & 20.97 & 12.73 & 12.31\\
            & HYDRA       & 61.04 & 29.90 & 25.55 & 25.04 & 24.11 & 22.36 & 18.73 \\
            & \name (ours)& 61.58 & \textbf{36.39} & \textbf{33.09} & \textbf{32.50} & \textbf{28.29} & \textbf{26.46}& 17.05 \\
        \hline
        \hline
        \multirow{3}{*}{16$\times$}
            & AdvPrune    & \textbf{68.24} & 38.98 & 23.20 & 20.50 & 19.13 & 8.40 & 12.08\\
            & HYDRA       & 61.35 & 29.14 & 25.53 & 24.85 & 23.92 & 21.95 & 18.77\\
            & \name (ours)& 61.84 & \textbf{35.78} & \textbf{32.24} & \textbf{31.34} & \textbf{27.31} & \textbf{25.28}& 17.09 \\
        \hline
    \end{tabular}}
\end{table}

\noindent\textbf{Pruning Rate Impact.}
We further measure the \switch{\arxiv}{performance}{natural accuracy and  robustness} of our \name and SOTA methods against all five attacks under 4$\times$, 8$\times$, and 16$\times$ pruning rate. We report these at 20\% mix ratio, so that training times are roughly equal across methods, \switch{\arxiv}{in}{in \Cref{tab:compare-cifar10} for CIFAR-10 and} \Cref{tab:compare-cifar100} for CIFAR-100. \emph{Overall, we can clearly see that \name consistently outperforms other SOTA methods against all five attacks, \stratis{under similar (or lower) training time}.} \tong{Specifically, \switch{\arxiv}{}{on CIFAR-100,} \name maintains high robustness against AA  with only 1.62\% drop (under 4$\times$ PR) from the pre-trained model by LBGAT (see \Cref{tab:pretrained}), while the AA robustness achieved by HYDRA and AdvPrune drop by 6.12\% and 13.98\%, respectively. This again verifies that, when pruning a robust model pre-trained with different adversarial training methods, \name is more stable in preserving robustness.
Improvements are also pronounced while increasing pruning rate:} \name outperforms HYDRA against AA by 4.50\%, 4.10\%, and 3.33\% under 4$\times$, 8$\times$, and 16$\times$ pruning rates, respectively. For completeness, we also report performance at 0\% mix ratio on CIFAR-100 in \switch{\arxiv}{\cite{pwoa-extend}}{Appendix~\ref{sec:appendix-mix0}}; in contrast to \name, competitors exhibit virtually negligible robustness in this case.

\switch{\arxiv}{}{
\begin{table}[!t]
    \centering
    \setlength{\tabcolsep}{2pt}
    \small
    \caption{\textbf{Comparison with APD.} Comparison \name with APD results reported in \cite{lee2021kdprune}, for pruning ResNet-18 on CIFAR-10 under 4$\times$ pruning rate. The authors report natural accuracy,  robustness under PGD$^{10}$, and number of epochs. We estimate execution time (T) per epoch and training time (TT, in \textit{h}), by training KD alone over adv.~examples.} 
    \label{tab:apd}
    \resizebox{0.45\textwidth}{!}{
    \begin{threeparttable}
        \begin{tabular}{||c |c|| c c || c c c||}
            \hline
            PR & Methods & Natural & PGD$^{10}$ & Epochs & T/epoch & TT\\
            \hline
            \hline
            \multirow{2}{*}{4$\times$} 
                & APD & 86.73\tnote{$\ast$} & $45.61^{\ast}$ & $60^{\ast}$ & $147.55s^{\dagger}$  & $2.46h{^\dagger}$ \\
                &\name & 86.07 & 49.61 & 150 & 45.19s & 1.88h \\
            \hline
        \end{tabular}
        \begin{tablenotes}
        {\scriptsize \item[*]reported in \cite{lee2021kdprune}. $^{\dagger}$estimated by KD over adv. examples.} 
        \end{tablenotes}
    \end{threeparttable}}
\end{table}

\noindent\textbf{Comparison with APD.}
Finally, we also compare to APD \cite{lee2021kdprune}, which is weaker than HYDRA and AdvPrune, but more closely related to our \name: APD prunes by optimizing KD \textit{over adversarial examples} using \textit{a non-robust teacher}.  \Cref{tab:apd} compares \name with APD on CIFAR-10 by ResNet-18 under 4$\times$ pruning rate (which is the largest pruning rate reported in their paper). We observe that, while achieving similar accuracy, \name outperforms APD w.r.t.~both robustness and training efficiency. This is expected, as distilling from a non-robust teacher limits APD's learning ability from adversarial examples and generating adversarial examples hampers training efficiency.}

\section{Conclusions and Future Work} \label{sec:conclusions}
We proposed \name, a unified framework for pruning adversarially robust networks without adversarial examples. Our method leverages pre-trained adversarially robust models,  preserves  adversarial robustness via self-distillation and enhances it via the Hilbert-Schmidt independence criterion as a regularizer. Comprehensive experiments on MNIST, CIFAR-10, and CIFAR-100 datasets demonstrate that \name prunes a large fraction of weights while attaining comparable adversarial robustness with up to 7$\times$ training speed up. Future directions include extending \name framework to structured pruning and weight quantization. Another interesting future direction is to use distillation and novel penalties to prune a pre-trained robust model even without access to natural examples.

\section{Acknowledgements}
The authors gratefully acknowledge support by the National Science Foundation under grants CCF-1937500 and CNS-2112471.


\switch{\arxiv}{}{\begin{appendices}
\section{Solving Problem (\ref{prob:prune-kd-hbar}) by ADMM}\label{sec:appendix-admm}

We follow \cite{zhang2018systematic,ren2019admm,Jian2021RFonEdge} in how to solve problem (\ref{prob:prune-kd-hbar} via ADMM. We begin by rewriting problem~(\ref{prob:prune-kd-hbar}) in the ADMM form by introducing auxiliary variables $\aux_l$:
\begin{equation}
    \begin{split} \label{prob:admm}
    &\underset{\bftheta}{\text{Minimize:}}\quad \nameloss(\bftheta)+\textstyle\sum_{l=1}^\nol g_l(\aux_l),\\
    &\text{subject to\quad} \bftheta_l = \aux_l, \quad l = 1, \cdots, \nol,
    \end{split}
\end{equation}
where $g_l(\cdot)$ is the indicator of set $S_l$, defined as:
{\begin{equation}
    g(\aux_l) = \left\{ 
    \begin{array}{ll}
    0&\text{if }\aux_l\in S_{l},\\
    +\infty&\text{otherwise.}\\
    \end{array}
    \right.
\end{equation}}

The augmented Lagrangian of problem  (\ref{prob:admm}) is \cite{boyd2011distributed}:
\begin{equation}\label{prob:lag}
\begin{split}
  \mathcal{L}&(\bftheta, \aux,\dual )= \nameloss(\bftheta) + \textstyle\sum_{l=1}^{\nol}g_l(\aux_l) \\
  &+\textstyle\sum_{l=1}^{\nol} \rho_l (\dual_l^\top (\bftheta_l-\aux_l ) )+ 
  \textstyle\sum_{l=1}^{\nol} \frac{\rho_{l}}{2}   \|{\bftheta}_{l}-\aux_{l} \|_{2}^{2}, 
  \end{split}
\end{equation}
where ${\rho_{l}}$ is a penalty value and ${\dual}_{l}\in\reals^{d_{\bftheta_l}}$ is a dual variable, rescaled by $\rho_l$.
The ADMM algorithm proceeds by repeating the following iterative optimization process until convergence. At the $k$-th iteration, the steps are given by 
\begin{subequations}\label{eq:admm-process}
    \begin{align}\label{eq:primal}
    \bftheta^{(k)} &:= \underset{\bftheta}{\argmin}\, \mathcal{L}(\bftheta,\aux^{(k-1)},\dual^{(k-1)})\\
   \label{eq:proximal}
    \aux^{(k)} &:= \underset{\aux}{\argmin}\,\mathcal{L}(\bftheta^{(k)}, \aux,\dual^{(k-1)})
  \\
  \label{eq:updateu}
    \dual^{(k)} &:= \dual^{(k-1)} + \bftheta^{(k)}-\aux^{(k)}.
    \end{align}
\end{subequations}
The problem \eqref{eq:primal} is equivalent to:  
\begin{equation} \label{eq:primal1}
    \min_{\bftheta}
    \nameloss(\bftheta) + \admmloss(\bftheta), 
\end{equation}
where 
\begin{equation} \label{eq:admmloss}
\admmloss(\bftheta)=\textstyle\sum_{l=1}^{\nol} 
    \frac{\rho_{l}}{2}\| \bftheta_{l} - \aux^{(k-1)}_{l}+\dual_{l}^{(k-1)}\|_F^2.
\end{equation}
All two terms in \eqref{eq:primal1} are quadratic and differentiable. Thus, this subproblem can be solved by classic Stochastic Gradient Descent (SGD).  
After solving problem \eqref{eq:primal} at iteration $k$, we proceed to solving problem \eqref{eq:proximal}, which is equivalent to:
\begin{equation} \label{eq:proximal1}
     \min_{\aux}
     \textstyle\sum_{l=1}^{\nol} g(\aux_{l})+\sum_{l=1}^{\nol}\frac{\rho_l}{2}\| \bftheta_l^{(k)}-\aux_{l}+\dual_{l}^{(k-1)}\|_F^2.
\end{equation}

As $g(\cdot)$ is the indicator function of the constraint set $S_l$, problem \eqref{eq:proximal1} is equivalent to:
\begin{equation} \label{eq:proximal2}
    \aux^{(k)}_l = \textstyle\proj_{S_{l}}\big(\bftheta_l^{(k)}+\dual_{l}^{(k-1)}\big),
\end{equation}
where $\proj_{S_{l}}$ is the Euclidean projection of $\bftheta_l^{(k)}+\dual_l^{(k-1)}$ onto the set $S_l$. 
The projection can be computed in polynomial time by first calculating  
$ \bftheta_l^{(k)}+\dual_l^{(k-1)}$, 
then keeping the $\alpha$ largest coefficients, in absolute value, and setting the rest to zero. 
%
The parameters $\bftheta$ produced by ADMM satisfy the constraints $\{S_l\}_{l=1}^\nol$ asymptotically. As a result, the fine-tuning process is typically required to improve the accuracy/robustness of the pruned model with the training dataset and attain feasibility~\cite{li2016pruning,wen2016learning,zhu2018improving, Ye2019advcompression}. To fine-tune the pruned model, we can construct a binary mask strictly satisfying the sparsity constraints $\{S_l\}_{l=1}^\nol$ and zero out weights that have been masked during back propagation. Formally, a binary mask is defined as $\bm{M}_l \in S_l \cap \{0,1\}^{d_{\bftheta_l}}$ for each layer $l$. 
The mask $\bm{M}_l$ is constructed as follows for irregular pruning: first, we compute $\bar{\aux_l}=\proj_{S_{l}}\left(\bftheta_l\right), l \in \{1,...,\nol\}$; then, we set $[\bm{M}_l]=\mathbf{1} \text{, for all entries s.t. } [\bar{\bftheta_l}]\neq 0 $. We then retrain $\bftheta$ using gradient descent but constrained by masks $\{\bm{M}_l\}_{l=1}^\nol$. That is, during back propagation, we first calculate the gradient $\nabla_{\bftheta_l} \nameloss(\bftheta_l)$, then apply the mask $\bm{M}_l$ to the gradient using element-wise multiplication. Therefore, the weight update in every step during the retraining process is 
\begin{equation}
\textstyle\bftheta_l  := \bftheta_l - \beta \bm{M}_l \circ \nabla_{\bftheta_l} \nameloss(\bftheta_l),
\end{equation}
where $\beta$ is the learning rate, and $\circ$ denotes element-wise multiplication.

\begin{table}[!t]
    \centering
    \setlength{\extrarowheight}{.2em}
    \setlength{\tabcolsep}{5pt}
    \small
    \caption{Parameter Summary. }
    \vspace{2pt}
    \label{tab:params}
    \resizebox{0.48\textwidth}{!}{
    \begin{tabular}{||c c || c | c | c||}
        \hline
        \multirow{2}{*}{Stage} & \multirow{2}{*}{Param.} & \multicolumn{3}{c||}{Dataset} \\
        \cline{3-5}
        && MNIST & CIFAR-10 & CIFAR-100 \\
        \hline
        \hline
        \multirow{7}{*}{Overall}
        & Batch size & 128 & 128 & 128 \\
        & Optimizer  & SGD & SGD & SGD \\
        & Scheduler  & cosine & cosine & cosine \\
        & $\tau$ ($\kdloss$) & 30 & 30 & 30 \\
        & $\lambda$ ($\kdloss$) & 10 & 10 & 1000 \\
        & $\lambda_x$ ($\hbarloss$) & 4e-4 & 2e-5 & 5e-7  \\
        & $\lambda_y$ ($\hbarloss$) & 1e-4 & 1e-4 & 2.5e-6 \\
        \hline
        \multirow{2}{*}{ADMM}
        & \# epochs  & 50 & 50 & 50 \\
        &Learning rate & 0.0005 & 0.01 & 0.01 \\
        
        \hline
        \multirow{2}{*}{Fine-tuning}
        &\# epochs   & 20 & 100 & 100 \\
        &Learning rate & 0.001 & 0.005 & 0.005  \\
        
        \hline
    \end{tabular}}
\end{table}

\section{Implementation Details}\label{sec:appendix-params}
We report the parameter settings in Table \ref{tab:params}. 

\noindent\textbf{ADMM Hyperparameters.} In pruning stage, we run ADMM every 3 iterations (Eq.~\eqref{eq:admm-process}). In each iteration, step \eqref{eq:primal} is implemented by one epoch of SGD over the dataset, solving Eq.~\eqref{eq:primal1} approximately. We set all $\rho_i=0.01$ initially; every iteration of ADMM, we multiply them by a factor of 1.35, until they reach 1. At the fine-tuning stage, we retrain the network under a pruned mask for several epochs.

\noindent\textbf{KD and HBaR Hyperparameters.} For $\kdloss$, we fix $\tau = 30$ in our experiments as we find that further tuning it leads to no performance gain. For $\hbarloss$, we follow original authors \cite{wang2021hbar} and apply Gaussian kernels for $X$ and $Z$ and a linear kernel for $Y$. For Gaussian kernels, we set $\sigma=5\sqrt{d}$, where $d$ is the dimension of the corresponding random variable. 

\noindent\textbf{\name Hyperparameters.} Recall that $\lambda$, $\lambda_x$ and $\lambda_y$ are balancing hyper-parameters for $\kdloss$ and $\hbarloss$, respectively. We first describe how to set $\lambda$: first, we compute the  value of $\kdloss$, $\hbarloss$ and $\admmloss$,  given by \eqref{prob:prune-kd}, \eqref{eq:obj} and \eqref{eq:admmloss} respectively, at the end of the first epoch. Then, we  set $\lambda$ so that   $\frac{\admmloss}{\lambda\kdloss}=10$; we empirically found that this ratio gives the best performance. 
Then, given this $\lambda$, we set $\lambda_x$ and $\lambda_y$ as follows. We follow Wang et al. \cite{wang2021hbar} to determine the ratio between $\lambda_x$ and $\lambda_y$: they suggest that setting the ratio $\lambda_x:\lambda_y$ as $4:1$ on MNIST, and as $1:5$ on CIFAR-10/100 provides better performance. We adopt these ratios, and scale both $\lambda_x$ and $\lambda_y$ (maintaning these ratios constant) so that $\frac{\lambda\kdloss}{\hbarloss}=10$; our choice of this ratio is determined empirically, by exploring different options. 

\tong{\noindent\textbf{Repetitions with Different Seeds.} Note that  \emph{initial weights are fixed in our setting}: we start from the pre-trained model, and repetition of experiments with different starting points does not apply to our setting. The only source of randomness  comes from (a) the SGD data sampler across epochs (b) and the adversarial example generation (in the mixed setting). Since we span 150 epochs, both processes are sampled considerably and our results are thus statistically significant.}

\section{Synergy between \hb Terms}\label{sec:appendix-hbar}
Figure~\ref{fig:hsic_plain} provides the learning dynamics on the HSIC plane for all datasets under 4$\times$ pruning rate. The x-axis plots HSIC between the last intermediate layer $Z_L$ and the input $X$, while the y-axis plots HSIC between $Z_L$ and the output $Y$. As discussed in \Cref{sec:prune-kd-hbar}, minimizing $\HSIC(X, Z_L)$ corresponds to reducing the influence of adversarial attack, while maximizing $\HSIC(Y, Z_l)$ encourages the discriminative nature of the classifier. The performance of different schemes can be clearly verified and demonstrated on the HSIC plain: as shown in Figure~\ref{fig:hsic_plain},  \name terminates with considerably lower $\HSIC(X, Z_L)$ than $\celoss$, indicating the stronger robustness against attacks.
Additionally, we observe the two optimization phases, especially on MNIST, separated by the start of fine-tuning stage: the \emph{risky compression phase}, where the top priority of the neural network is to prune non-important weights while maintain meaningful representation by increasing $\HSIC(Y, Z_L)$ regardless of the information redundancy ($\HSIC(X, Z_L)$), and the \emph{robustness recovery phase}, where the neural network turns its focus onto inheriting robustness by minimizing $\HSIC(X, Z_L)$, while keeping highly label-related information for natural accuracy.

\begin{figure}[!t]
   \centering
   \small
   \includegraphics[width=0.85\columnwidth]{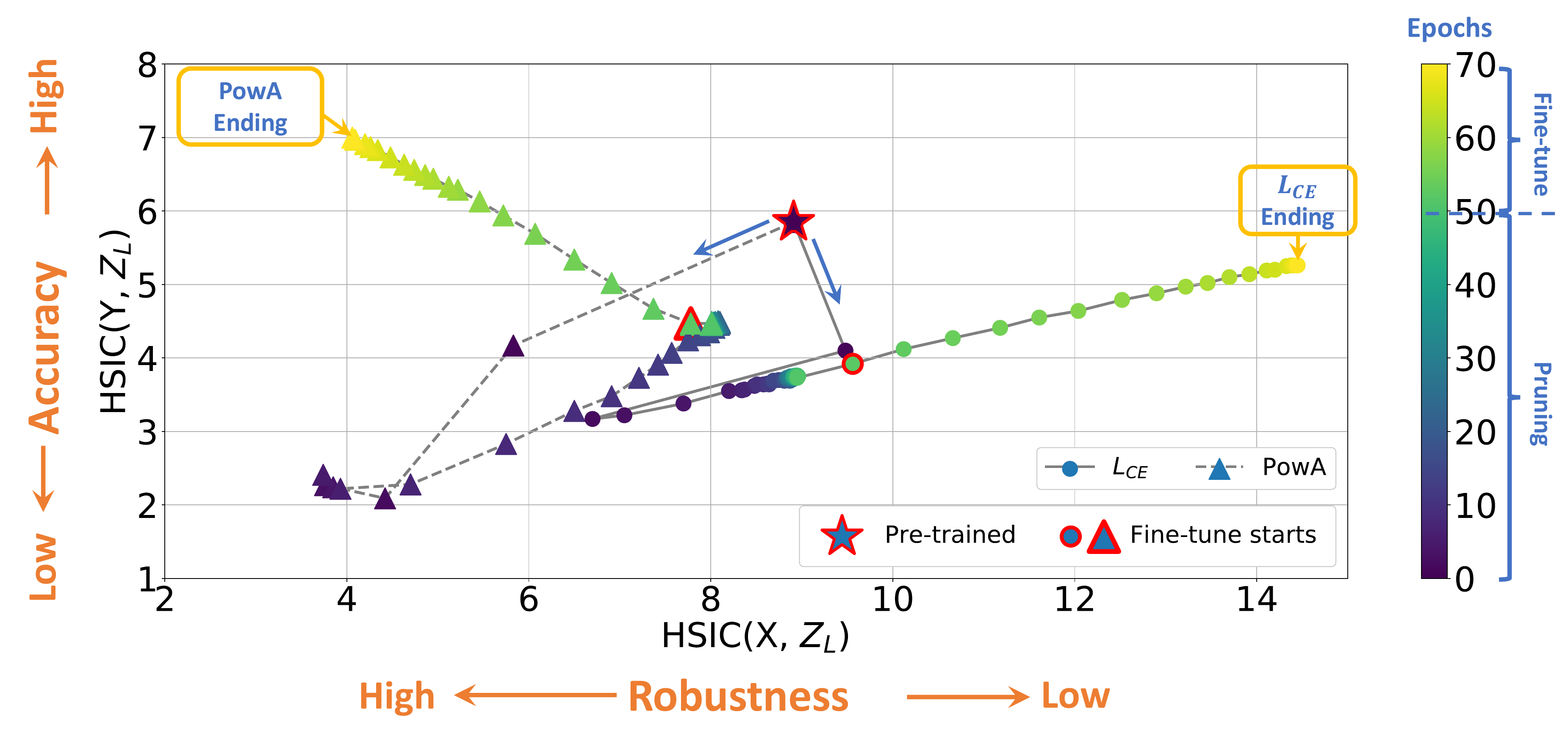} \\
	(a) MNIST by ResNet-18 \\
   \includegraphics[width=0.85\columnwidth]{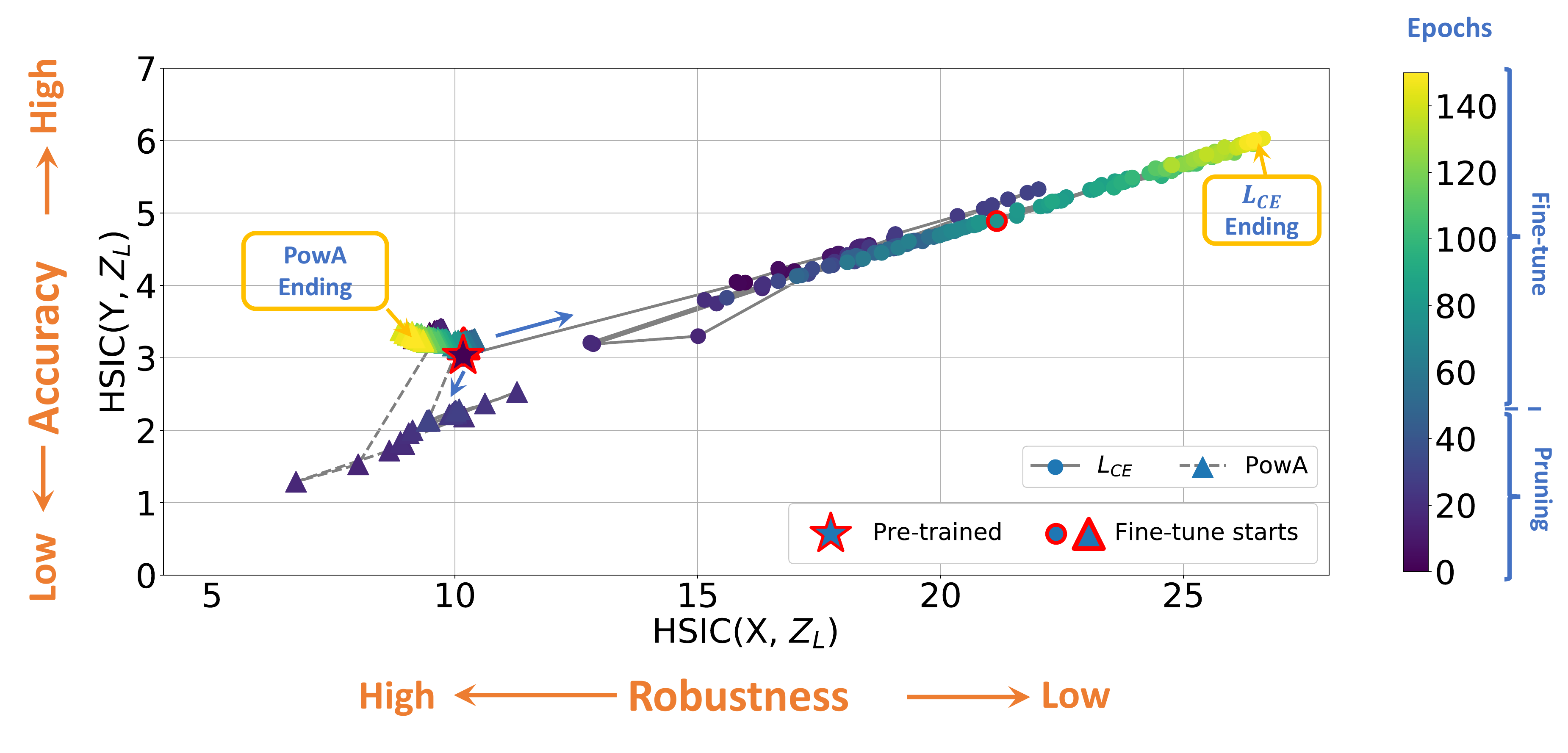} \\
	(b) CIFAR-10 by WRN34-10 \\
	\includegraphics[width=0.85\columnwidth]{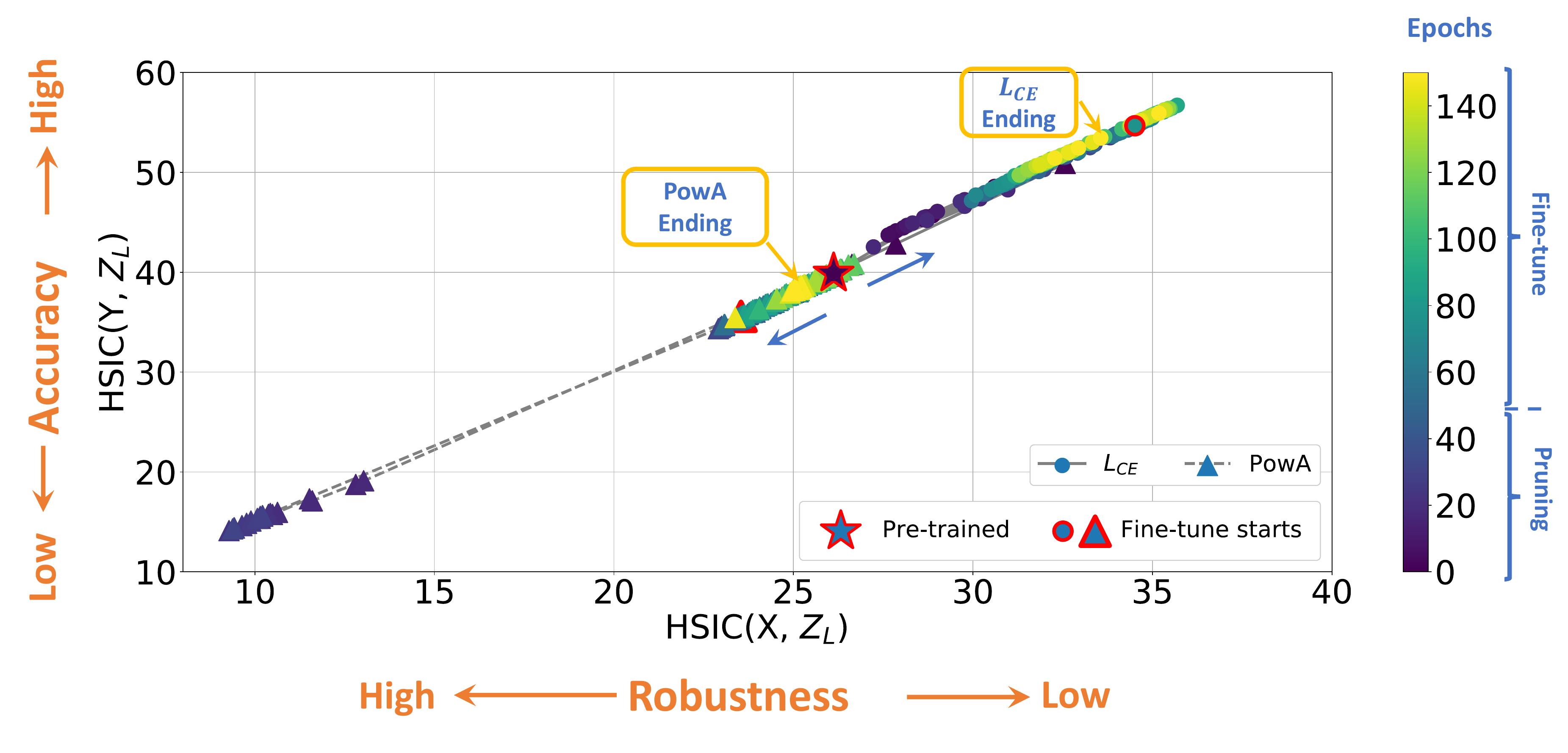} \\
	(c) CIFAR-100 by WRN34-10 \\
	\caption{HSIC plane dynamics. The x-axis plots HSIC between the last intermediate layer $Z_L$ and the input $X$, while the y-axis plots HSIC between $Z_L$ and the output $Y$. The color scale and arrows indicate indicate dynamic direction w.r.t. training epochs. Each marker in the figures represents a different setting: \textbf{stars}, \textbf{dots}, and \textbf{triangles} represent pre-trained, $\celoss$, and \name, respectively.}
	\label{fig:hsic_plain}
\end{figure}

\begin{table}[!t]
    \centering
    \setlength{\extrarowheight}{.2em}
    \setlength{\tabcolsep}{3pt}
    \small
    \caption{\textbf{Prune WRN34-10 (LBGAT) on CIFAR-100}: Comparison of \name with SOTA methods w.r.t various attacks and training time (TT, in \textit{h}) under different pruning rates at 0\% mix ratio.}
    \vspace{2pt}
    \label{tab:compare-cifar100-mix0}
    \resizebox{0.48\textwidth}{!}{
    \begin{tabular}{||c |c || c c c c c c| c |}
        \hline
        PR & Methods & Natural & FGSM & PGD$^{10}$ & PGD$^{20}$ & CW & AA & TT \\
        \hline
        \hline
        \multirow{3}{*}{4$\times$}
            & AdvPrune    & 79.56 & 17.31 & 0.24 & 0.10 & 0.03 & 0.00 & 2.62 \\
            & HYDRA       & 79.62 & 17.39 & 0.25 & 0.10 & 0.03 & 0.00 & 5.70 \\
            & \name (ours)& 60.92 &	36.70 &	33.08 &	32.59 &	28.4 & 26.44 & 9.33  \\
        \hline
        \hline
        \multirow{3}{*}{8$\times$}
            & AdvPrune    & 79.38 & 16.98 & 0.18 & 0.08 & 0.02 & 0.00 & 2.64 \\
            & HYDRA       & 79.44 & 17.21 & 0.21 & 0.10 & 0.02 & 0.00 & 5.54 \\
            & \name (ours)& 61.43 &	35.61 &	31.19 &	30.45 &	26.32 &	24.20 & 9.32 \\
        \hline
        \hline
        \multirow{3}{*}{16$\times$} 
            & AdvPrune    & 79.21 & 16.98 & 0.18 & 0.08 & 0.02 & 0.00 & 2.61 \\
            & HYDRA       & 79.36 & 17.18 & 0.20 & 0.09 & 0.02 & 0.00 & 5.79\\
            & \name (ours)& 62.53 &	35.15 &	29.05 &	27.88 &	24.08 &	21.43 & 9.18 \\
        \hline
    \end{tabular}}
\end{table}

\section{Pruning Rate Impact at 0\% Mix Ratio}\label{sec:appendix-mix0}
We evaluate natural accuracy and the robustness of our PwoA and SOTA methods against all five attacks under 4$\times$,  8$\times$, and 16$\times$ pruning rate as well as the overall training time, and report these at 0\% mix ratio in \Cref{tab:compare-cifar100-mix0} for CIFAR-100. We observe that, without access to adversarial examples (mix ratio 0\%), both competing methods fail catastrophically, exhibiting no robustness whatsoever. Moreover, increasing the pruning rate can lead to sharp drop in robustness, especially with limited access to adversarial examples. Not surprisingly, comparing to \Cref{tab:compare-cifar100} at 20\% mix ratio, this drop occurs much severer in method being more dependent on adversarial learning objectives, e.g., under 8$\times$ against AA, HYDRA drops from 22.26\% (at 20\% mix ratio) to 0.00\% (at 0\% mix ratio) while \name drops from 26.46\% to 24.20\%; under 16$\times$ against AA, HYDRA drops from 21.95\% to 0.00\% while \name drops from 25.28\% to 21.43\%.

\end{appendices}

}
\end{document}